\documentclass[manuscript]{acmart}
\AtBeginDocument{%
  }

\setcopyright{acmlicensed}
\copyrightyear{2026}
\acmYear{2026}
\acmDOI{XXXXXXX.XXXXXXX}
\acmConference[TOMM]{Transactions on Multimedia Computing Communications and Applications}{June 03--05, 2026}{Woodstock, NY}
\acmISBN{978-1-4503-XXXX-X/18/06}

\usepackage{makecell} 
\usepackage[normalem]{ulem} 
\usepackage{xcolor} 
\usepackage{multirow} 
\usepackage{soul}

\newcommand{\revise}[2]{{#2}}

\newcommand{\new}[1]{{#1}}

\begin{document}

\title{A Survey on Deep Learning Architectures for Point Cloud Classification and Segmentation}

\author{Minhas Kamal}
\email{mxkamal@albany.edu}
\orcid{0009-0008-4514-7846}
\affiliation{%
  \institution{State University of New York at Albany}
  \city{Albany}
  \state{New York}
  \country{USA}
}

\author{Hiranya Garbha Kumar}
\email{hgkumar@albany.edu}
\orcid{0000-0003-1493-3840}
\affiliation{%
  \institution{State University of New York at Albany}
  \city{Albany}
  \state{New York}
  \country{USA}
}

\author{Balakrishnan Prabhakaran}
\email{bprabhakaran@albany.edu}
\orcid{0000-0003-0385-8662}
\affiliation{%
  \institution{State University of New York at Albany}
  \city{Albany}
  \state{New York}
  \country{USA}
}

\renewcommand{\shortauthors}{Minhas Kamal et al.}

\setcctype{by}
\acmJournal{TOMM}
\acmYear{2026} \acmVolume{1} \acmNumber{1} \acmArticle{}
\acmMonth{1} \acmDOI{10.1145/3815180}

\begin{abstract}

Point cloud stands as the most widely adopted format for representing 3D shapes and scenes due to its simplicity and geometric fidelity. However, its inherent unordered and irregular nature, exacerbated by sensor noise and occlusions, introduces unique challenges for machine learning based methodologies. To combat these issues, diverse strategies have been developed, including converting to a format that has orderliness, extracting local geometry, and permutation-invariant or self-attention-based processing. In this paper, our focus is directed towards deep learning models for three fundamental tasks in 3D vision: point cloud classification, part segmentation, and semantic segmentation. We begin by formally defining point cloud data, followed by an in-depth discussion on its structural characteristics. Then, we categorize notable works based on their backbone structure and evaluate their performance on popular benchmarks. Beyond empirical comparison, we offer insights into architectural innovations and limitations. We also outline open challenges and promising future directions for 3D point cloud understanding.


\end{abstract}



\received{18 Jul 2025}
\received[revised]{13 Dec 2025}
\received[accepted]{29 Apr 2026}

\maketitle

\section{Introduction}
\label{sec:introduction}

Point cloud processing has emerged as a cornerstone of 3D vision and recent developments in self-supervised learning, attention mechanisms, state-space models, and multi-modal frameworks have significantly advanced the field. Nevertheless, critical challenges such as scalability, robustness, and cross-domain generalization persist, calling for continued research into efficient and adaptable learning architectures tailored for 3D point cloud understanding.

\subsection{Background}
\label{subsec:background}

For over six decades, computer vision has remained one of the most prominent and actively researched domains within the field of artificial intelligence. In contrast, research on understanding 3D geometric data did not experience significant traction until the early 2010s, partially due to the scarcity of large-scale, high-quality annotated datasets, as well as the lack of cost-effective and readily available sensor technology \cite{Wu153DShapeNets, Guo21Deep-dl3dsurvey}. The absence of efficient algorithms and the substantial computational resources required for processing 3D data further impeded the progress in this field.

Nonetheless, the range of potential applications and commercial opportunities for 3D data processing continues to expand. Key domains include autonomous vehicles and robotics, geographic information systems (GIS), augmented and virtual reality (AR/VR), medical imaging, digital gaming, computer-aided design and manufacturing (CAD/CAM), and archaeological reconstruction. Consequently, with the emergence of commodity 2.5D cameras such as- Microsoft Kinect (2010) and Intel RealSense (2014), and affordable LiDAR (Light Detection and Ranging) technologies such as- Velodyne VLP-16 (2016) and Livox Mid-40 (2019), the interest of the research community got renewed for 3D data analysis. 

\subsection{Motivation}
\label{subsec:motivation}

The introduction of affordable and accessible 3D sensors led to the creation of large-scale benchmark datasets like- KITTI (2012) \cite{Geiger12KITTI}, S3DIS (2016) \cite{Armeni16S3DIS}, ScanNet (2017) \cite{Dai17Scannet}, and \revise{ScanObjectNN (2019) \cite{Uy19ScanObjectNN}}{Waymo Open Dataset (2020) \cite{Sun20WaymoOpenDataset}}, and open source libraries like- Point Cloud Library (2011) \cite{Rusu11PCL}, Open3D (2018) \cite{Zhou2018Open3D}, and PyTorch3D (2020) \cite{Ravi20Pytorch3D}. Advancements in computational resources like graphics processing unit (GPU) and pioneering algorithmic innovations like- \revise{RANSAC (1981) \cite{Fischler81RANSAC}, }{}COLMAP (2016) \cite{Schonberger16COLMAP}, PointNet (2017) \cite{Charles17PointNet}, \revise{DGCNN (2019) \cite{Wang19dgcnn}, Minkowski Engine (2019) \cite{Choy19MinkowskiEngine}, }{}Point Cloud Transformer (2021) \cite{Guo21PCT}, Neural Radiance Fields (2021) \cite{Mildenhall21NeRF}, 3D Gaussian Splatting (2023) \cite{kerbl233DGS}, and Depth Anything (2024) \cite{Yang24DepthAnything}, have only expanded the potential use cases across numerous disciplines.

In recent years, 3D computer vision has become one of the most prolific research domains in artificial intelligence, with thousands of peer-reviewed papers published annually across top-tier venues such as the IEEE/CVF Conference on Computer Vision and Pattern Recognition (CVPR), the International Conference on Computer Vision (ICCV), the European Conference on Computer Vision (ECCV), as well as broader AI-focused conferences including the Conference on Neural Information Processing Systems (NeurIPS), the International Conference on Machine Learning (ICML), and the International Conference on Learning Representations (ICLR). In addition, leading journals such as the International Journal of Computer Vision (IJCV), the IEEE Transactions on Pattern Analysis and Machine Intelligence (TPAMI), and Nature Machine Intelligence regularly publish influential contributions in the field. For instance, more than 15\% of all papers featured at \revise{ICCV '23}{CVPR '25} addressed challenges within the domain of 3D data processing. This surge in research output reflects the growing academic and industrial interest in 3D data understanding and its diverse applications.

\subsection{Goal and Structure}
\label{subsec:goal-and-structure}


The development of state-of-the-art (SOTA) techniques for 3D data processing has progressed in lockstep with advances in deep learning (DL) \cite{LeCun15DeepLearning}. Since the advent of artificial neural networks, DL has matured into a foundation for landmark computer-vision architectures, including the Multi-Layer Perceptron (MLP, 1986) \cite{Rumelhart86MLP}, Convolutional Neural Networks (CNNs, 1989) \cite{LeCun89CNN, Long15Fully}, diffusion models (2015) \cite{SohlDickstein15Diffussion, Ho20DDPM}, and the Vision Transformer (ViT, 2021) \cite{Vaswani17Transformer, Dosovitskiy21ViT}. Although originally designed for 2D images, successive variants of these models have been adapted to ingest 3D shapes. Nevertheless, recent SOTA performance gains in 3D processing techniques appear to stem primarily from improved training protocols, such as sophisticated data-augmentation and optimization strategies, and increased model capacity, rather than from fundamental architectural innovations, as highlighted by PointNeXt \cite{Qian22PointNext}.


This survey offers a targeted examination of DL approaches to 3D point cloud data (PCD) processing, with a particular focus on the fundamental tasks of shape classification, semantic segmentation, and part segmentation. We also review the principal public datasets and highlight their role in helping address the distinctive challenges posed by point clouds. Although several earlier surveys have covered similar ground \cite{Zhang19Review-dl3dsurvey, Bello20Review-dl3dsurvey, Xie20Linking-dl3dsurvey, Guo21Deep-dl3dsurvey, Zhang23DeepLearningBased, Sarker24Comprehensive-dl3dsurvey}, many are now outdated due to the rapid progress spurred by recent advances in DL. Moreover, recent comprehensive reviews tend to catalog datasets and benchmark results without critically evaluating architectural choices or their practical trade-offs. In contrast, our review adopts a selective, in-depth perspective, concentrating on landmark contributions that have substantively advanced 3D computer vision. Rather than compiling exhaustive tables, we probe the theoretical foundations and empirical performance of key backbone architectures, dissecting their strengths, limitations, and design rationales. We conclude with analytical insights and prospective research directions, furnishing both newcomers and experienced researchers with a thorough, critical view of the current state and future potential of DL for point cloud analysis.

\section{Point Cloud}
\label{sec:point-cloud}

Point cloud data depicts the surface geometry of an object as a set of discrete data points defined in Cartesian coordinates. Each point has spatial coordinates and may include additional attributes such as color or normal. Although point clouds can be used for any dimensional surface representation, here, we are only considering 3D space. So, a point cloud can be mathematically represented as a set of 3D points:

\begin{equation}
\mathcal{P} = \{ \mathbf{p}_i \in \mathbb{R}^3 \mid i = 1, 2, \dots, N \}, \quad \mathbf{p}_i = (x_i, y_i, z_i) ^ \top
\label{eq:point-cloud}
\end{equation}

$\mathbf{p}_i$ represents the 3D coordinates of the $i$-th point in Euclidean space. Additional attributes such as color components $(r_i, g_i, b_i)$ and surface normals $(n_i^x, n_i^y, n_i^z)$ can also be included to the point:

\begin{equation}
\mathbf{p}_i = (x_i, y_i, z_i, r_i, g_i, b_i, n_i^x, n_i^y, n_i^z, \ldots) ^ \top
\label{eq:point}
\end{equation}

\revise{Point cloud can be easily derived by sampling points from the surface of a mesh or by fusing multiple depth maps through methods like SLAM (Simultaneous Localization and Mapping). As there is no explicit connectivity information among the points, point cloud does not directly represent any surface. Also, a point cloud is unordered and unstructured. Moreover, real-world point clouds contain lots of artifacts and occlusions. However, point cloud is simple and is arguably the most popular modality for recording, representing, and processing 3D shapes.}{A point cloud can be derived by sampling mesh surfaces or by fusing multiple depth maps. As there is no explicit connectivity information among the points, it does not directly represent any surface and is inherently unordered and unstructured. Moreover, real-world point clouds contain lots of artifacts and occlusions. However, a point cloud is simple and is arguably the most popular modality for recording, representing, and processing 3D shapes.}

\begin{figure}[ht]
    \centering
    \includegraphics[width=\textwidth]{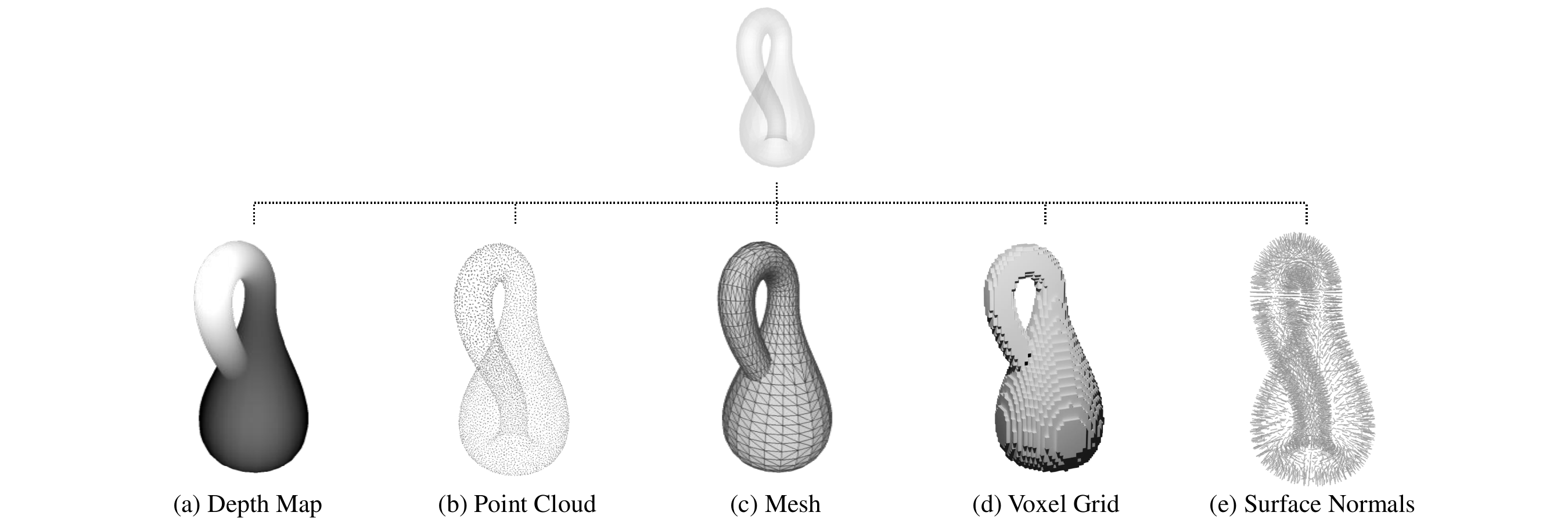}
    \caption{Common 3D scene representation modalities illustrated by a Klein bottle.}
    \label{fig:3d-representation}
    \Description{3d-representation}
\end{figure}

There are several other widely recognized 3D geometric data representation techniques (shown in Figure \ref{fig:3d-representation}), each designed to fulfill a specific set of requirements. Appendix \ref{sec:other-3d-structure-representation-modalities} provides a short overview of each of \revise{the other common 3D data representation modalities}{them}.

\subsection{Point Cloud Datasets}
\label{subsec:datasets}

Compared to the widespread availability and diverse use cases of 2D images, 3D data remains relatively less adopted and presents greater complexity in acquisition. Also, with the rise and widespread usage of social media, the volume of 2D visual content has exploded, driven by the ease of capturing, sharing, and consuming photos and videos using smartphones. This proliferation has led to significant advances in 2D computer vision supported by massive datasets. In contrast, the growth of 3D visual data has not followed a similar trajectory. As a result, large-scale and high-quality 3D datasets are rare and have become a significant bottleneck for training and evaluating machine learning (ML) models.

\begin{figure}[ht]
    \centering
    \includegraphics[width=\textwidth]{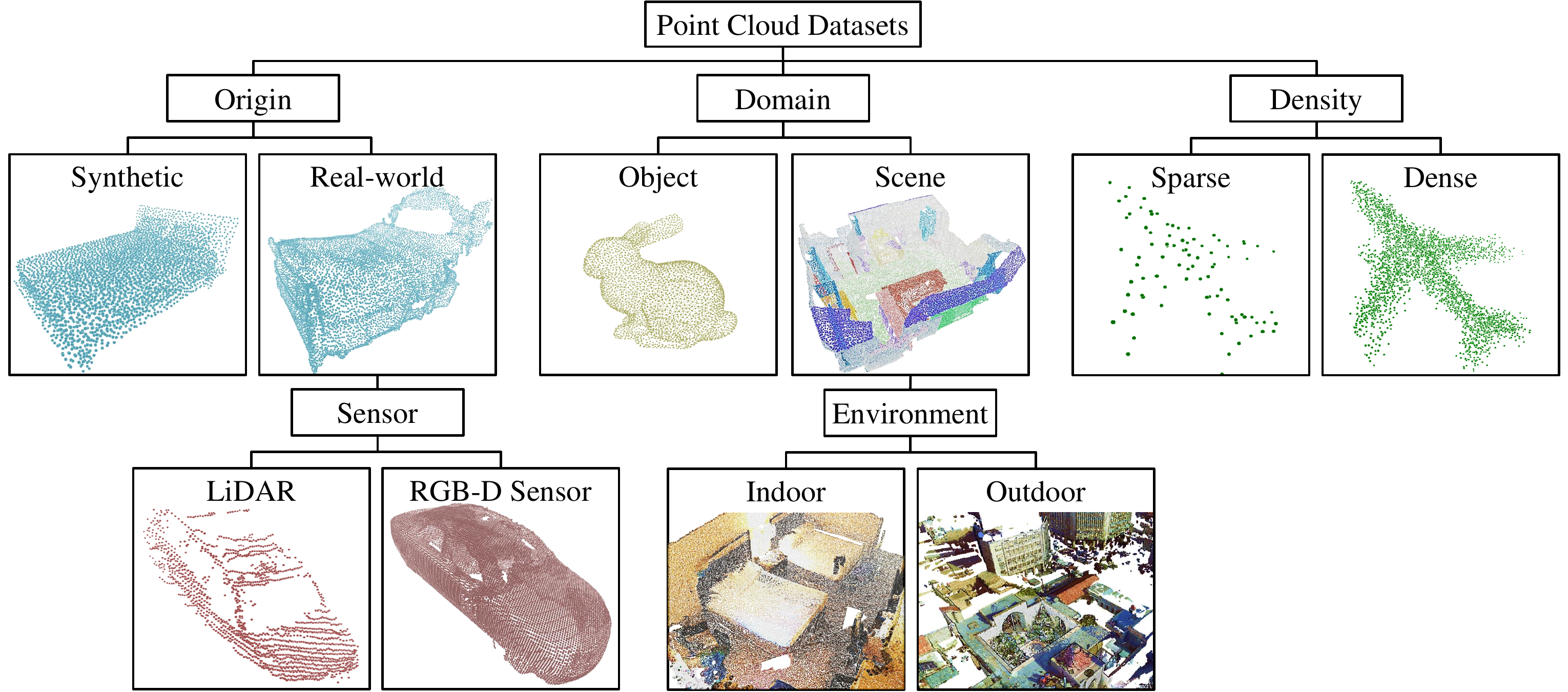}
    \caption{\revise{Various ways of classifying point cloud datasets.}{Classification of point cloud datasets from multiple perspectives.}}
    \label{fig:3d-dataset-classification}
    \Description{3d-dataset-classification}
\end{figure}

Most of the 3D datasets are in the form of RGB-D images, meshes, or point clouds. Point clouds can be directly derived from RGB-D data using methods such as SLAM (Simultaneous Localization and Mapping), or from meshes by sampling points on their surfaces, enabling a unified representation across modalities. Figure \ref{fig:3d-dataset-classification} illustrates the classification of 3D datasets from three different points of view.

Based on data origin, 3D datasets can be categorized as \textbf{synthetic} or \textbf{real-world}. Synthetic datasets, such as ModelNet \cite{Wu153DShapeNets} and ShapeNet \cite{Chang15Shapenet}, consist of CAD models that are hand-crafted by humans and represented as 3D meshes. Synthetic datasets are simpler to deal with because there are no occlusions or noise associated with them. In contrast, Real-world datasets, such as KITTI \cite{Geiger13KITTI}, ScanNet \cite{Dai17Scannet}, and nuScenes \cite{Caesar20nuScenes}, are acquired by scanning the real world with 3D sensors and often contain artifacts. There are primarily two types of 3D sensors: LiDAR and RGB-D. Outdoor scenes (KITTI, nuScenes, etc.) are usually scanned by LiDARs, and indoor scenes (ScanNet, ScanObjectNN \cite{Uy19ScanObjectNN}, etc.) by RGB-D cameras.

Datasets also differ in scope: some point cloud datasets contain the whole scene (ScanNet, S3DIS \cite{Armeni16S3DIS}, etc.), while some only contain individual objects (ScanObjectNN, ShapeNet, etc.). Finally, depending on sensor characteristics or sampling strategies, point clouds may vary in density, ranging from sparse to dense representations. COLMAP \cite{Schonberger16COLMAP} generates a sparse point cloud by estimating camera poses and 3D structure via Structure from Motion (SfM) from multi-view image collections. See Figure \ref{fig:3d-dataset-classification} for a visual illustration of the different categories of point cloud datasets.

\renewcommand{\arraystretch}{1.5}
\renewcommand\cellgape{\Gape[4pt]}

\begin{table}
 \caption{Representative point cloud datasets. Abbreviations: \textit{sc}- scanned scenes, \textit{sca}- individual scans, \textit{cls}- object classes, \textit{n}- samples, \textit{img}- images.}
\label{tab:point-cloud-datasets}
\resizebox{\textwidth}{!}{
\begin{tabular}{cccccccc}
  \toprule
  \multirow{2}{*}{\textbf{Dataset}}
  & \multirow{2}{*}{\textbf{Year ↑}}
  & \multirow{2}{*}{\textbf{Type}}
  & \multirow{2}{*}{\textbf{Size}}
  & \multicolumn{4}{c}{\textbf{Benchmark Task}} \\
  \cline{5-8}
  &
  &
  &
  & \textbf{\rotatebox[origin=c]{75}{Classification}}
  & \textbf{\rotatebox[origin=c]{75}{Segmentation}}
  & \textbf{\rotatebox[origin=c]{75}{Detection}}
  & \textbf{\rotatebox[origin=c]{75}{Reconstruction}}\\
  \midrule
  KITTI \cite{Geiger12KITTI} & 2012 & LiDAR, Outdoor (Driving), Sparse & sc=22, cls=8, n=15k & - & \checkmark & \checkmark & - \\
  NYU Depth V2 \cite{Silberman12NYUDepthV2} & 2012 & RGB-D, Indoor, Dense & sc=464, cls=14, n=1.5k & - & \checkmark & - & - \\
  \makecell{Sydney Urban \\ Objects \cite{Deuge13SydneyUrbanObjects}} & 2013 & LiDAR, Outdoor (Driving), Sparse & cls=14, n=588 & \checkmark & - & - & - \\
  ModelNet \cite{Wu153DShapeNets} & 2015 & Synthetic, Single Object & cls=40, n=12k & \checkmark & - & - & - \\
  ShapeNet \cite{Chang15Shapenet} & 2015 & Synthetic, Single Object & cls=55, n=50k & \checkmark & - & - & \checkmark \\
  SUN RGB-D \cite{Song15SUNRGBD} & 2015 & RGB-D, Indoor, Dense & sc=47, cls=37, n=58k, img=10k & \checkmark & \checkmark & \checkmark & \checkmark \\
  S3DIS \cite{Armeni16S3DIS} & 2016 & RGB-D, Indoor, Dense & \makecell{sca=272, sc=6, cls=12 (+1 clutter), \\ points=215M} & - & \checkmark & - & - \\
  SceneNN \cite{Binh16SceneNN} & 2016 & RGB-D, Indoor, Dense & sc=101, cls=40 & - & \checkmark & - & - \\
  Semantic3D \cite{Hackel17Semantic3D} & 2017 & LiDAR, Outdoor, Dense & sc=30, cls=8 & - & \checkmark & - & - \\
  ScanNet \cite{Dai17Scannet} & 2017 & RGB-D, Indoor, Dense & \makecell{sca=1513, cls=17 (20 semantic),\\ n=12k, img=2.5M} & \checkmark & \checkmark & \checkmark & - \\
  Matterport3D \cite{Chang17Matterport3D} & 2017 & RGB-D, Indoor, Dense & sc=90, cls=40, img=200k & - & \checkmark & - & - \\
  Pix3D \cite{Sun18Pix3d} & 2018 & Synthetic+RGB, Indoor & cls=9, n=10k (395 unique) & \checkmark & - & - & \checkmark \\
  ScanObjectNN \cite{Uy19ScanObjectNN} & 2019 & RGB-D, Single Object, Dense & cls=15, n=15k (3k unique) & \checkmark & - & - & - \\
  nuScenes \cite{Caesar20nuScenes} & 2020 & LiDAR, Outdoor (Driving), Sparse & sc=1k, cls=23, n=40k, img=1.4M & - & \checkmark & \checkmark & - \\
  \new{\makecell{Waymo Open \\Dataset} \cite{Sun20WaymoOpenDataset}} & 2020 & LiDAR, Outdoor (Driving), Sparse & sc=1150, cls=4, n=230k & - & \checkmark & \checkmark & - \\
  Clear Pose \cite{Avidan22ClearPose} & 2022 & RGB-D, Indoor, Dense & cls=63, n=350k & - & - & \checkmark & - \\
  \bottomrule
  \end{tabular}
}
\end{table}

Table \ref{tab:point-cloud-datasets} offers a comparative overview of popular benchmark datasets for point clouds, Appendix \ref{sec:acquisition} outlines the sensors and techniques used for sourcing these datasets, and Appendix \ref{sec:datasets} provides a short description of them. For performance evaluation of the classification models, we shall utilize results on the ModelNet40 \cite{Wu153DShapeNets} and the ScanObjectNN \cite{Uy19ScanObjectNN} dataset. ShapeNet-Part \cite{Chang15Shapenet} and S3DIS \cite{Armeni16S3DIS} datasets are employed for comparing the results of part segmentation and semantic segmentation methods, respectively.

\subsection{Point Cloud Processing Tasks}
\label{subsec:point-cloud-processing-tasks}

Point cloud processing encompasses a broad range of tasks which also includes classical computer vision tasks: \textbf{filtering}- denoising and/or outlier removal to improve data fidelity; \textbf{classification}- assigning semantic label to a point cloud object; \textbf{object detection}- identifying and localizing instances within a 3D scene; and various forms of segmentation, such as: \textbf{part segmentation}- decomposing objects or scenes into constituent parts; \textbf{semantic segmentation}- labeling each point with a semantic class; and \textbf{instance segmentation}- similar to semantic segmentation but also distinguishing between individual object instances of the same class. \revise{As point cloud processing is computationally expensive, \textbf{point cloud sampling} is a vital preprocessing step in 3D data pipelines, significantly influencing the performance of downstream tasks. Therefore, a good sample should be lightweight and representative of the shape.}{As point cloud processing is computationally expensive, \textbf{point cloud sampling} serves as a vital preprocessing step. The quality of the sampled point cloud significantly influences the performance of downstream tasks, making it essential to be both lightweight and representative of the shape.}

Some of the general 3D data manipulation tasks also require point cloud processing, like converting to other data representation modalities: \textbf{voxelization}- transforming unstructured point clouds into structured volumetric representations; \textbf{mesh generation} and \textbf{normal estimation}- estimating 3D surface and its orientation by connecting neighboring points. Some common inter point cloud tasks are: \textbf{registration}- aligning multiple point clouds into a unified coordinate space; \textbf{upsampling}- leveraging generative models to increase the density and resolution of a point cloud; and \textbf{generation}- generating 3D point cloud by conditioning on a text prompt or a 2D image. \textbf{3D face recognition}, \textbf{human skeleton detection}, \textbf{pose understanding}, and \textbf{action recognition} are some other tasks that also involve point cloud processing.

Beyond these, there are specialized tasks in point cloud processing, like reconstructing a 3D scene: \textbf{dense reconstruction}- synthesizing 3D geometry by combining a series of scans; \textbf{monocular 3D reconstruction}- estimating depth, orientation, and occluded parts from a single or multiple 2D images to generate corresponding 3D structures; and \textbf{semantic reconstruction}- integrates geometry and semantic information to produce a labeled and interactive 3D scene. Reconstruction task might involve several sub-tasks: \textbf{completion}- inferring missing regions of partially observed 3D data; \textbf{pose estimation}- determining the spatial orientation (rotation, translation, and scale) of 3D entities; \textbf{CAD retrieval}- matching point clouds to a database of known 3D models; and \textbf{CAD deformation}- stretching and shearing CAD models to match a certain geometric shape.

\begin{figure}[ht]
    \centering
    \includegraphics[width=0.70\textwidth]{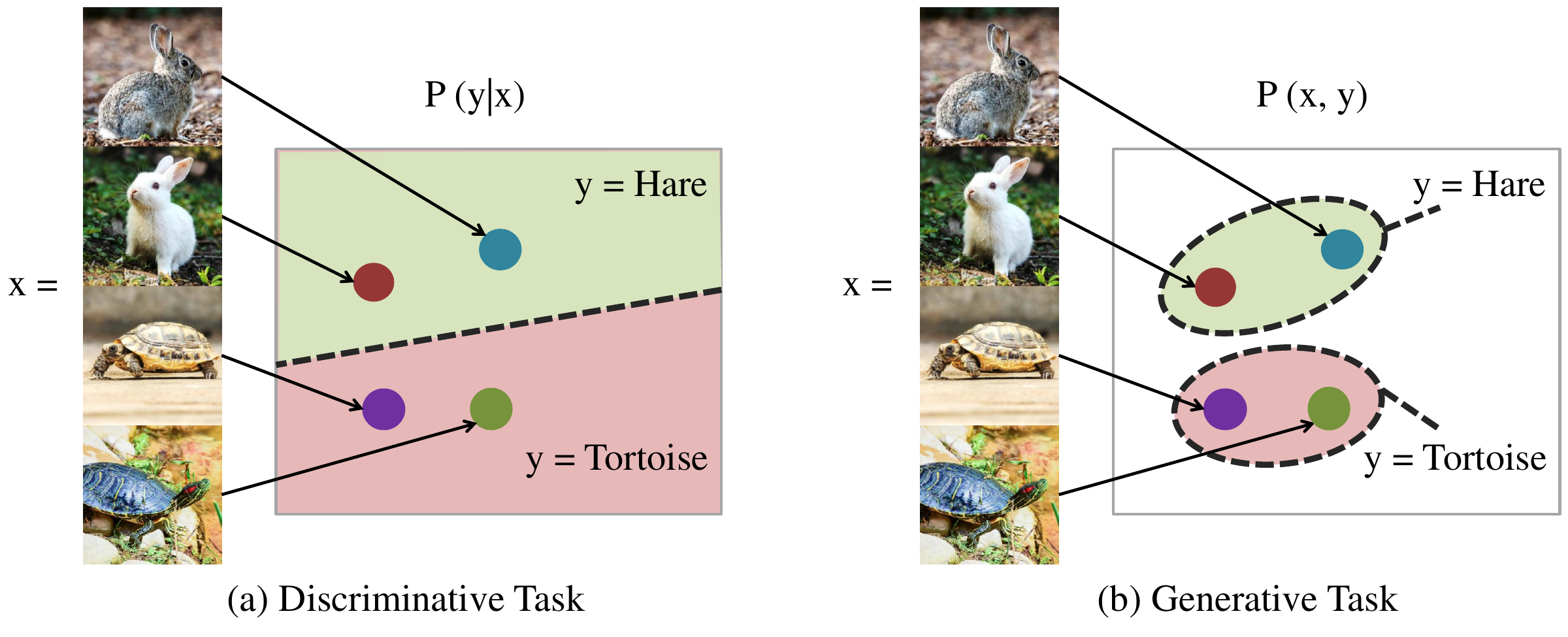}
    \caption{\revise{Demonstrates a comparative illustration of discriminative and generative tasks.}{{Comparative illustration of discriminative and generative modeling paradigms.}} (a) Discriminative models learn the conditional distribution- $P(y|x)$. (b) Generative models learn the joint distribution- $P(x, y)$.}
    \label{fig:discriminative-and-generative-tasks}
    \Description{discriminative-and-generative-tasks}
\end{figure}

All the aforementioned vision tasks can be broadly categorized into two principal groups, as illustrated in figure \ref{fig:discriminative-and-generative-tasks}, based on the underlying learning objectives: (1) \textbf{Discriminative Tasks}- models are trained to infer decision boundaries and distinguish between class distributions, examples include classification, semantic/instance segmentation, and pose estimation; and (2) \textbf{Generative Tasks}- models aim to learn the underlying data distribution itself, such as in point cloud upsampling, shape completion, and 3D object generation. Generative modeling typically presents greater challenges due to the complexity of capturing the full variability and structure inherent in the data distribution. And so, a model that is trained on generative tasks can be repurposed for discriminative tasks too, but the opposite is not possible.

\subsection{Opportunities and Challenges with Point Cloud}
\label{subsec:opportunities-and-challenges-with-point-cloud}

\subsubsection{Opportunities}
\label{subsubsec:opportunities}
Among the different formats for 3D structure representation, point cloud has emerged as the most prevalent format for contemporary 3D ML tasks, including classification and segmentation \cite{Bello20Review-dl3dsurvey, Xie20Linking-dl3dsurvey, Guo21Deep-dl3dsurvey, Sarker24Comprehensive-dl3dsurvey}. This widespread adoption is largely attributed to several inherent conveniences that the point cloud format offers, which are effectively leveraged by ML methodologies for 3D shape analysis.

\begin{figure}[ht]
    \centering
    \includegraphics[width=\textwidth]{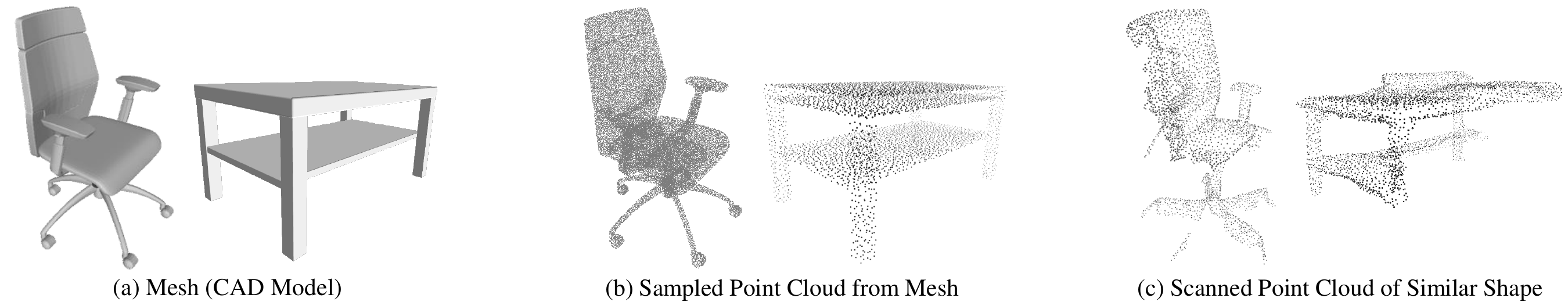}
    \caption{Point clouds are unordered, irregular, random, and do not represent any surface. As a result, even the synthetically generated perfect point clouds (b) are difficult to process. Real-world point clouds (c) introduce more challenges- noise, occlusions, artifacts, and background clutter.}
    \label{fig:pointcloud-probs}
    \Description{pointcloud-probs}
\end{figure}

\paragraph{Mathematical Simplicity and Extensibility}
A 3D point cloud is mathematically defined as an unordered set of points, where each point is represented by a triplet (x,y,z) corresponding to its spatial coordinates in Euclidean space. So simply, a point cloud is a set of coordinates. Moreover, it is readily extensible: each point can also have auxiliary attributes such as RGB color values, surface normals, reflectance intensity, or semantic labels. This enables point cloud data to support a wide range of 3D perception tasks.

\paragraph{Sensor Fidelity and Direct Acquisition}
Point cloud is the closest to raw sensor data compared to other 3D structure representation formats, as it can be derived directly from LiDAR and depth sensors. So, it would be much efficient to process point cloud directly instead of converting it to an intermediary data such as voxel or render 2d images, and then processing this intermediate data. Plus, the conversion step might lose valuable information gathered by the sensors as well as propagate errors.

\paragraph{Ease of Derivation from Other 3D Formats}
We can sample point clouds directly from polygonal meshes and voxel grids. The sampling algorithm can be controlled to satisfy different requirements- random or uniform, dense or sparse, partial or complete, and so on. This facilitates the integration of heterogeneous 3D geometric data sources within a unified processing pipeline and conducts comparative evaluations.

\paragraph{Scalability and Efficiency}
point cloud representations scale well to large and complex scenes, and can represent fine geometric structures without high memory cost. It is achieved through adaptive geometric sampling- higher point density is used where surface variation is significant, such as edges and corners. This reduced memory footprint usually interprets to lower computational overhead for ML models. In addition, point clouds from multiple viewpoints can be trivially combined to form a more comprehensive representation of objects or scenes.

\subsubsection{Challenges}
\label{subsubsec:challenges}
Figure \ref{fig:pointcloud-probs} helps to realize the key challenges in processing point clouds by DL architectures. Some of these challenges are intrinsic to the data format, while others stem from limitations and inefficiencies in contemporary 3D sensing technologies and algorithms.

\paragraph{Unordered}
The primary challenge in processing point clouds lies in their lack of inherent order or regularity, in contrast to other data types such as tables, text, RGB images, videos, voxels, and depth maps. For example, altering the position of a pixel in an image or a word in a sentence changes the information conveyed. However, in a point cloud, the position of a point carries no semantic meaning. As a result, conventional neural network architectures, such as CNNs \cite{LeCun89CNN}, Deep Belief Networks (DBNs) \cite{Hinton06AFast}, or MLPs \cite{Rumelhart86MLP}, are not suitable for directly processing 3D point clouds. Instead, a network architecture that is invariant to the ordering of points—typically utilizing a symmetric function—must be employed. This architecture should also be capable of effectively capturing the local structure of each individual point.

\paragraph{Lack of Surface and Texture Information}
Point clouds are discrete samples from an object’s surface and do not encode any explicit connectivity or topology. Each point exists independently, with no information about adjacency or surface continuity. So, there can be multiple objects placed side by side, and a point cloud does not contain any information about where the surface of one object ends and another begins. Although point clouds can contain color information, as there is no surface, there is a lack of surface-texture information.

\paragraph{Irregular and Non-Uniform Distribution}
Real-world point clouds are irregularly sampled and spatially non-uniform. Common issues include non-uniform point spacing, holes in data due to reflective or occluded surfaces, and varying resolution depending on object distance and material properties.

\paragraph{Unstructured and Random}
A 3D point cloud is simply a set of (x, y, z) coordinates, each indicating a point in 3D space. Unlike voxels or meshes, point clouds do not conform to a fixed spatial arrangement or graph structure. Furthermore, the stochastic acquisition process (e.g., in LiDAR scans) introduces randomness in point placement, which affects the reproducibility and stability of learning-based methods.

\paragraph{Presence of Noise and Artifacts}
Sensor noise is a pervasive issue in point cloud acquisition, arising from various sources including electronic interference, environmental lighting conditions, and surface reflectivity. Noise can manifest as spurious or ghost points, spatial jitter, or variation in point density, particularly in edge regions or across different material types. Also, there can be small objects attached to the surface of large objects, such as books inside a shelf of keys on a table. 

\paragraph{Incomplete and Occluded}
For capturing the complete 3D scene, we have to place the sensor in a higher dimension, just like we can capture a 2D plane from a 3D space with a regular RGB camera. As we exist in a 3D space, we cannot capture a complete 3D scene in reality. Instead, we can capture only a 2D projection of the 3D environment. Multi-view fusion strategies are used to reconstruct a more holistic representation from the projections. But even with careful acquisition attempts, occlusions and missing data remain prevalent in real-world datasets.

\paragraph{Low Resolution and Sparseness}
Point clouds, particularly in large-scale outdoor environments such as those encountered in autonomous driving, often suffer from inherent sparsity. Distant objects may be represented by a limited number of points, which impedes the ability to accurately infer their shape, texture, or semantic category. Furthermore, the low resolution of point cloud restricts the capacity to capture fine details, making it challenging to model intricate features in the environment. This sparsity and resolution limitation present significant obstacles for tasks that require precise and detailed 3D representations.

Despite the challenges associated with capturing and processing 3D point clouds, a complete 3D scan of a scene, typically obtained by integrating data from frames of an RGB-D video alongside information from Inertial Measurement Unit (IMU) sensors, remains the most reliable and comprehensive representation of a 3D environment. For this reason, it continues to be the preferred input method for research focused on holistic 3D scene understanding. \new{The subsequent sections will first examine prominent architectures for point cloud processing tasks, followed by the metrics commonly used to benchmark these methods.}


\section{Classification and Segmentation Architectures}
\label{sec:classification-and-segmentation-architectures}

Classification involves assigning probability values to each of the corresponding classes for a given piece of input. This probability value indicates the likelihood of the input data belonging to a specific class, such as animals, cars, furniture, etc. 

While classification assigns a class value to the entire point cloud, segmentation aims to provide a more granular understanding by allocating a class label to each individual point in the cloud. Among the different types of segmentation tasks, we shall focus on- semantic segmentation and part segmentation. Semantic segmentation \new{implies partitioning a scene into semantically meaningful regions by providing a semantic label for every point}; for example, decomposing a rural landscape into regions labeled as grass, water-body, house, tree, and sky. Part segmentation involves segmenting individual objects into different components; for example, dividing a motorcycle image into wheels, seats, oil tank, engine, etc. 

Most contemporary 3D instance segmentation approaches are built upon semantic segmentation frameworks, often extending them with additional clustering or grouping mechanisms to distinguish object instances. Therefore, instance segmentation is not treated as a separate topic in this review, given its architectural and conceptual overlap with semantic segmentation.

\begin{figure}[ht]
    \centering
    \includegraphics[width=\textwidth]{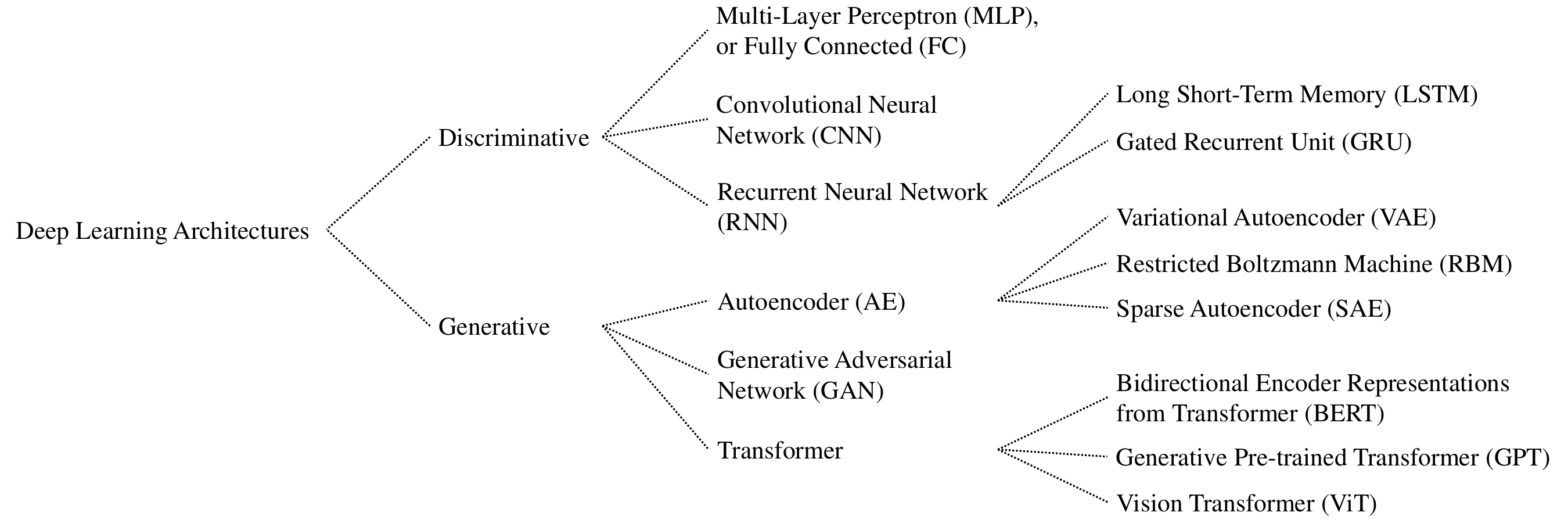}
    \caption{Taxonomy of commonly used deep learning architectures. \new{Note that architectural categories of discriminative and generative are not strictly exclusive, and some models can be adapted for both purposes. The classification shown here reflects their most common use cases.}}
    \label{fig:deeplearning-arch}
    \Description{deeplearning-arch}
\end{figure}

Figure \ref{fig:deeplearning-arch} demonstrates a comprehensive taxonomy of deep learning (DL) architectures spanning various domains beyond computer vision. Contemporary DL-based systems predominantly operate on raw point clouds, enabling end-to-end learning directly. However, in earlier approaches, point clouds were often transformed into more structured representations such as volumetric pixel (voxel) grids or multi-view 2D projections to facilitate compatibility with an adapted variant of conventional frameworks. These intermediate formats allowed the reuse of well-established frameworks such as 2D CNN, but often incurred significant information loss or computational overhead. Below, we will look into the architectural innovation behind each of these methodologies.

\begin{figure}[ht]
    \centering
    \includegraphics[width=\textwidth]{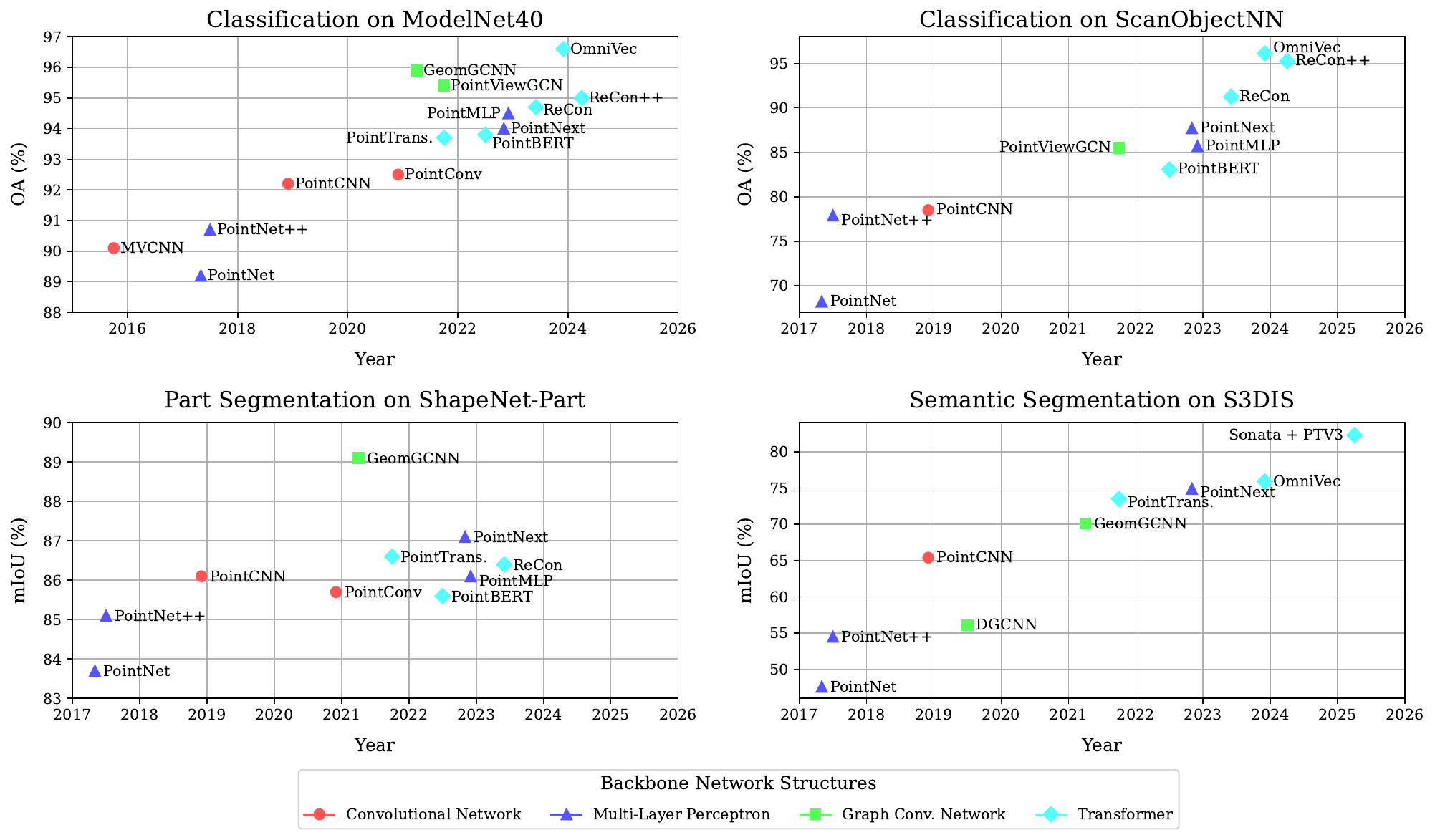}
    \caption{Comparison among different methods for classification and segmentation tasks.}
    \label{fig:model-comparison}
    \Description{model-comparison}
\end{figure}

\subsection{Convolutional Network}
\label{subsec:convolutional-network}

CNN has revolutionized computer vision tasks, because its architectural structure leverages the learning of the hierarchical features present in an image \cite{LeCun89CNN, Krizhevsky12AlexNet, Simonyan15VGG, Szegedy15GoogLeNet, He16ResNet}. Different varieties of CNNs have dominated SOTA lists of popular benchmarks for every category in vision learning tasks.

\begin{displaymath}
\text{Classical CNN Architecture: } (Conv,ReLU,MaxPool) \times n \rightarrow Flatten \rightarrow (FC,ReLU) \times m \rightarrow FC \rightarrow Softmax
\end{displaymath}

Early works on point cloud processing have been mostly on tailoring CNNs for different types of representations extracted from 3D structural data. Among the different approaches, the subsequent three are the most noteworthy: (1) Converting point clouds into a structured format, such as a voxel grid, and then running convolution with 3D kernels, (2) Rendering 2D images from the point cloud and processing through regular 2D CNNs, and (3) Adapting CNNs for ingesting unordered point clouds directly. Hence, depending on the type of data that the network directly ingests, we can divide the point cloud processing tasks into the following three categories.

\subsubsection{Volumetric Pixel}
\label{subsubsec:convolutional-network}

From the perspective of conventional 2D raster image representation, volumetric pixels (voxels) appear to be the most intuitive and natural extension for representing 3D data, offering a structured and regular grid analogous to 2D pixels. Consequently, several works have tried to process voxel grids directly using 3D CNNs- VoxNet (2015) \cite{Maturana15VoxNet}, 3DShapeNets (2015) \cite{Wu153DShapeNets}, OctNet (2017) \cite{Riegler17OctNet}, O-CNN (2017) \cite{Wang17OCNN}, VoxelNet (2018) \cite{Zhou18VoxelNet}, PointGrid (2018) \cite{Le18PointGrid}, among others. In these approaches, voxel grids are typically generated from raw point clouds or CAD models. In the following section, we examine the pipeline of 3DShapeNets, one of the earliest and most influential voxel-based architectures.

\begin{figure}[ht]
    \centering
    \includegraphics[width=\textwidth]{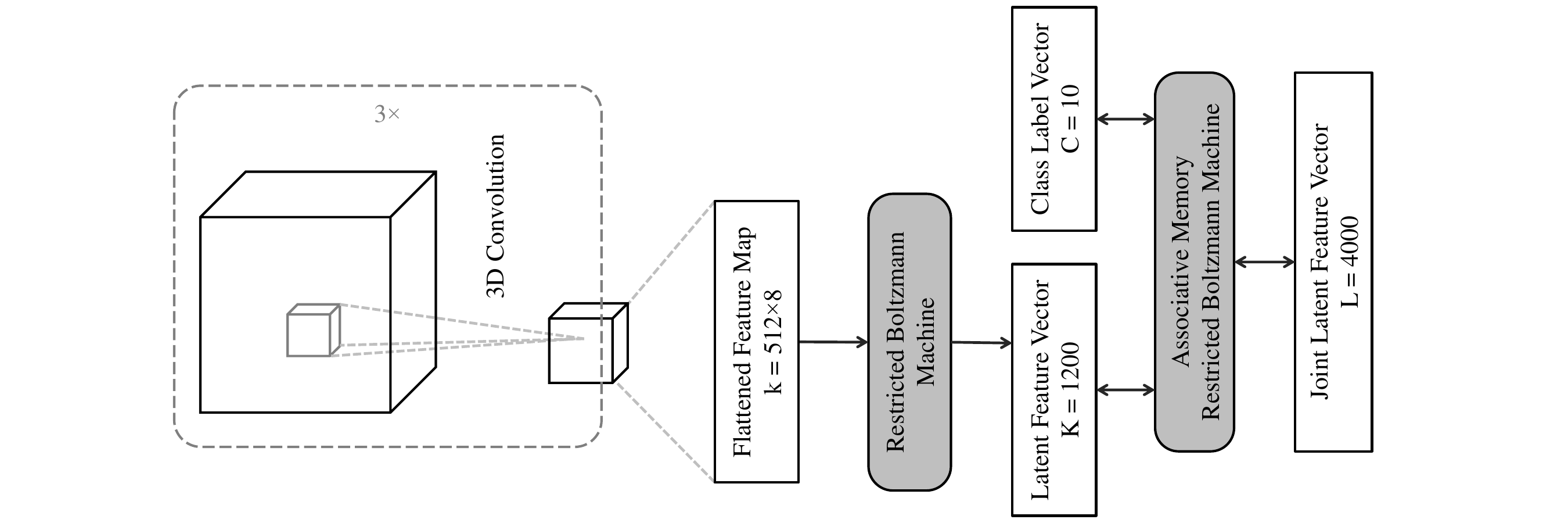}
    \caption{A simplified architecture of 3DShapeNets (2015) \cite{Wu153DShapeNets}. Spatial features are extracted from the voxel using 3 layers of 3D convolution, followed by a Restricted Boltzmann Machine (RBM) for global feature extraction of dimension $K$. A subsequent RBM jointly embeds the extracted features and class labels into a shared latent space for final classification.}
    \label{fig:3dshapenets-architecture}
    \Description{3dshapenets-architecture}
\end{figure}

\textbf{3DShapeNets} starts by taking a $(30\times30\times30)$ grid of voxels that contains the 3D shape as a $(24\times24\times24)$ grid with a padding of $(3+3)$. If a voxel in the grid is part of the 3D shape, then it has a value of 1, else the value is 0 (empty space). The authors use the network architecture of a convolutional deep belief network \cite{Hinton06AFast, Lee11Unsupervised} adapted from 2D pixel data to 3D voxel data and train the joint distribution $P(x,y)$ where $x$ is voxel data and $y$ is the object category label. Figure \ref{fig:3dshapenets-architecture}  provides a simplified depiction of 3DShapeNets.

First, they have 3 layers of 3D CNN and then 1 layer of Restricted Boltzmann Machine (RBM) and then 1 final layer of RBM modeled over Associative Memory Deep Belief Network (AM-DBN). Each CNN layer applies multiple 3D convolutional filters with a stride for extracting features from the shape. They do not use any pooling, as it was observed that pooling introduced uncertainty to shape reconstruction.

They pretrain the model first and then run fine-tuning. Pre-training is run layer-wise-convolution layers and RBM layer are trained with standard contrastive divergence \cite{Hinton02Training} and AM-DBN layer is trained with fast persistent contrastive divergence \cite{Tieleman09Using}. For fine-tuning, they use a process similar to the wake-sleep algorithm from \cite{Hinton06AFast}. During wake, they propagate input voxel forward through the network and update the recognition weights. During sleep, they sample persistent latent variables from the network's generative distribution and propagate them backward through the network to update the generation weights.

As high-resolution voxels are expensive to process, 3DShapeNets ingests shape information of only 24 voxels in each dimension. This coarse structure generalizes over the details of the source shape. In effect, the model produces only 77\% overall accuracy over ModelNet40 \cite{sun22ModelNet40C}, which does not have any noise or occlusion. OctNet \cite{Riegler17OctNet} occupies less memory and requires less runtime while working with higher resolution data, as it encodes volumetric information hierarchically in an octree. PointGrid \cite{Le18PointGrid} introduces a hybrid network that ingests both point and low resolution voxel grid for efficiently extracting geometric information using 3D-CNNs.

\subsubsection{Projected View}
\label{subsubsec:projected-view}

To combat the unavailability of large-scale 3D shape data and leverage robust network architectures developed for 2D images, several works have focused on working with projected views. More number of projections implies better shape depiction while increasing the processing overhead. So, researchers have experimented with various orientations for capturing the most descriptive projection while minimizing the count. The most renowned multi-view-based approaches are- MVCNN (2015) \cite{Su15MVCNN}, GVCNN (2018) \cite{Feng18GVCNN}, and View-GCN (2020) \cite{Wei20ViewGCN}. Below, we present a detailed overview of one of the pioneering and most widely adopted methods in this category- MVCNN.
 
\begin{figure}[ht]
    \centering
    \includegraphics[width=\textwidth]{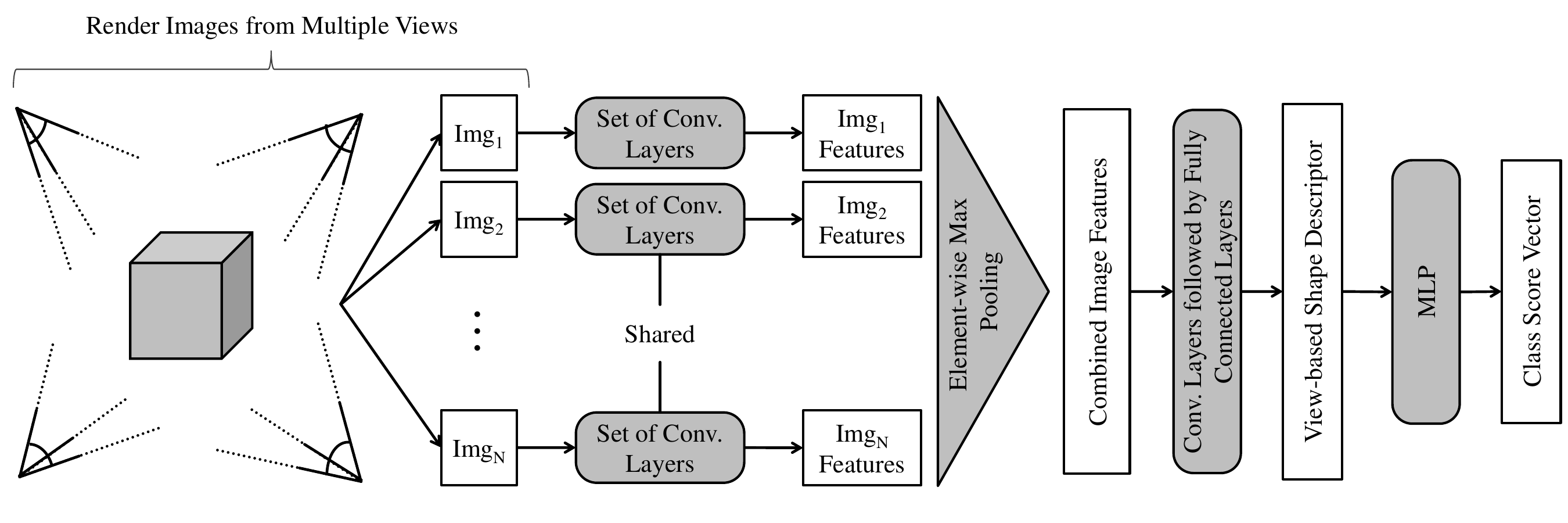}
    \caption{A simplified architecture of MVCNN (2015) \cite{Su15MVCNN}. Images taken from $N$ different views are individually processed by a set of shared convolution layers. These view-specific features are then aggregated by max pooling, and then processed by another convolution and a fully connected layer for global feature extraction.}
    \label{fig:mvcnn-architecture}
    \Description{mvcnn-architecture}
\end{figure}

The principal contribution of \textbf{MVCNN} is to build a view-based shape descriptor for 3D models where the descriptor offers a more efficient means for downstream tasks, like- classification and retrieval. First, the researchers scale the 3D shape uniformly to fit the viewing field and then render 80 projections of it from 20 viewpoints. They put the object in the middle of an icosahedron (a 3D shape with uniformly placed 20 vertices), place the camera in the positions of the vertices, and take 4 shots each with a 90\textdegree rotation of the shape.

Now, each rendered 2D image is individually passed through 4 convolution layers. This gives out features for each 2D image. These features from 80 images are then aggregated using a view-pooling layer. View-pooling layer runs element-wise max-pooling across all views. These combined features are then passed through a final convolution layer, followed by 2 fully connected layers. This outputs the aggregated multi-view-based shape descriptor that captures the holistic structure of the 3D shape albeit being compact and efficient to process. This shape descriptor is passed through a fully connected layer and a softmax layer for classification and retrieval tasks.

So altogether, there are 5 convolution layers and 3 fully connected layers, which follow the same architecture as the VGG-M network \cite{Chatfield14Return, Simonyan15VGG} with an additional view-pulling layer in the middle. The network is first pretrained with the ImageNet dataset \cite{Deng09Imagenet} and then fine-tuned using the 3D training data. We have compiled a simplified design of MVCNN in Figure \ref{fig:mvcnn-architecture}.

The advantage of view-based methods is that- they can directly implement the mature CNN architecture developed for 2D image processing. Also, they can use existing large-scale image datasets for pretraining. Being able to pretrain is important as it helps build a robust 3D data processing model even with the absence of domain-specific large-scale datasets. However, a view-based method cannot achieve the ability to completely represent a 3D structure thus the projection phase loses information. Furthermore, real-world data is often irregular, noisy, and includes background clutter and occlusions. As a result, projection-based techniques are generally only effective in controlled, synthetic environments and struggle to generalize to real-world scenarios.

\subsubsection{Point Set}
\label{subsubsec:point-set}

Both voxel-based and view-based methods offer ordered and regular representations of 3D shapes, allowing them to leverage well-established network architectures originally developed for 2D image processing. However, these methods are generally only effective in synthetic environments and fall short in performance compared to modern point set-based approaches. Voxel-based methods are limited in their ability to capture high-resolution geometric details due to their cubic memory growth with increasing resolution, which significantly raises computational costs. View-based methods, on the other hand, suffer from inherent information loss during the projection of 3D data onto 2D planes, limiting their fidelity. In contrast, point clouds represent the raw output of most 3D scanning sensors, making them a more direct and efficient input format. Converting point clouds to an intermediary format also introduces additional computational overhead. That is why, recent machine learning research has increasingly focused on developing methods that operate directly on point clouds. PointCNN (2018) \cite{Li18PointCNN}, PCNN (2018) \cite{Atzmon18PCNN}, PointConv (2019) \cite{Wu19PointConv}, and KPConv (2019) \cite{Thomas19KPConv} are some of the most prominent works that have revised convolution mechanisms for running directly on point clouds.


\begin{figure}[ht]
    \centering
    \includegraphics[width=0.96\textwidth]{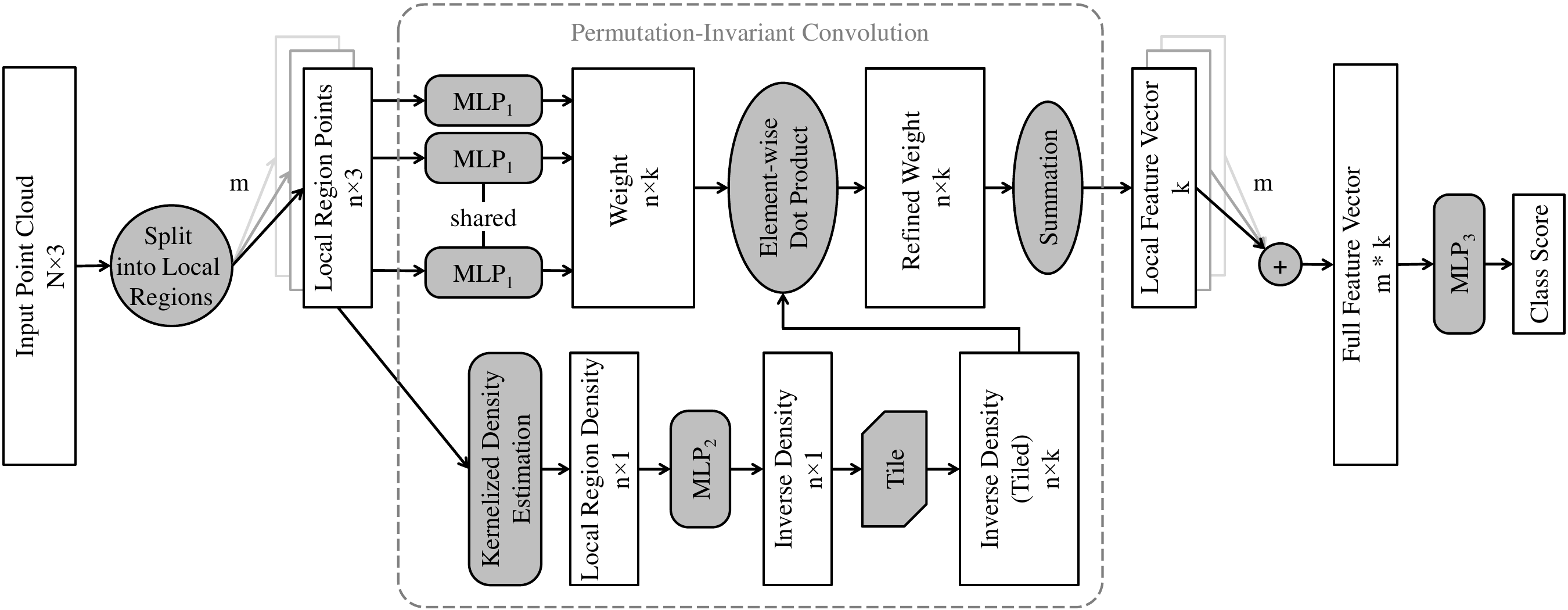}
    \caption{A simplified architecture of PointConv (2019) \cite{Wu19PointConv}. A point cloud of $N$ points is subdivided into $m$ regions, each with $n$ points. A Permutation-Invariant Convolution layer extracts $k$ features from each of these local point clouds. The features are later concatenated to be used for downstream tasks.}
    \label{fig:pointconv-architecture}
    \Description{pointconv-architecture}
\end{figure}

Observing the potential in CNN’s robustness and efficiency, researchers of \textbf{PointConv} \cite{Wu19PointConv} designed techniques for running permutation-invariant convolution on the point cloud directly. As point clouds are very different than 2D raster images, they have redefined the concept of local region, convolution kernel, and convolution operation, especially for point clouds. In a 3D point cloud, a local region is composed of a central point and a specific number of neighboring points in 3D space. A local region represents a local geometry, and the coordinates of the member points are updated to the local reference point. The convolutional kernels are defined as continuous convolution weight functions. These learned weight functions are convolved over each local region for calculating the weight or feature at a particular location. But before running convolution on a local region, it is first passed through an inverse density function, which eliminates the non-uniformity of the points while calculating the output weights.

For a given local region, the weight functions are learned using a shared MLP, and the inverse density scale is estimated through kernel density estimation followed by an MLP. This allows PointConv to run translation-invariant and permutation-invariant convolutions on local regions. Figure \ref{fig:pointconv-architecture} demonstrates the basic framework of PointConv. We can plug PointConv into any point cloud classification or segmentation network arrangement as the feature extractor.

\subsection{Multi-Layer Perceptron}
\label{subsec:multilayer-perceptron}

\textbf{PointNet} (2017) \cite{Charles17PointNet} is the pioneering deep network architecture that directly works on point clouds as input. It introduced a clever architecture that can directly process any unordered data. It produced competitive results in object classification and segmentation while competing with mature image processing networks, like- DBNs and CNNs, which consumed voxelized data \cite{Wu153DShapeNets, Riegler17OctNet, Wang17OCNN, Le18PointGrid} or rendered 2d images \cite{Su15MVCNN, Feng18GVCNN, Wei20ViewGCN}. The key idea behind the PointNet comes from the observation of the order-agnostic nature of symmetric functions over input data. Symmetric functions such as addition or max-operation or multiplication produce the same result no matter which permutation the variables are arranged.

\begin{equation}
f(x_1, x_2, ... , x_n) = \gamma [MAX_{i=1,2,...,n} \{h(x_i)\}]
\label{eq:pointnet}
\end{equation}

Function \ref{eq:pointnet} demonstrates the basic processes running in PointNet. Each point of the point cloud data goes through a function $h$ which encodes the coordinates to a higher dimension. Then, a max-pooling function pools the most prominent features from the points. This function is the heart of PointNet, which captures the holistic feature of the shape while not being affected by the order of the input. It enables the overall result to be permutation invariant. Function $\gamma$ is another nonlinear activation function for processing the shape feature.

\begin{figure}[ht]
    \centering
    \includegraphics[width=\textwidth]{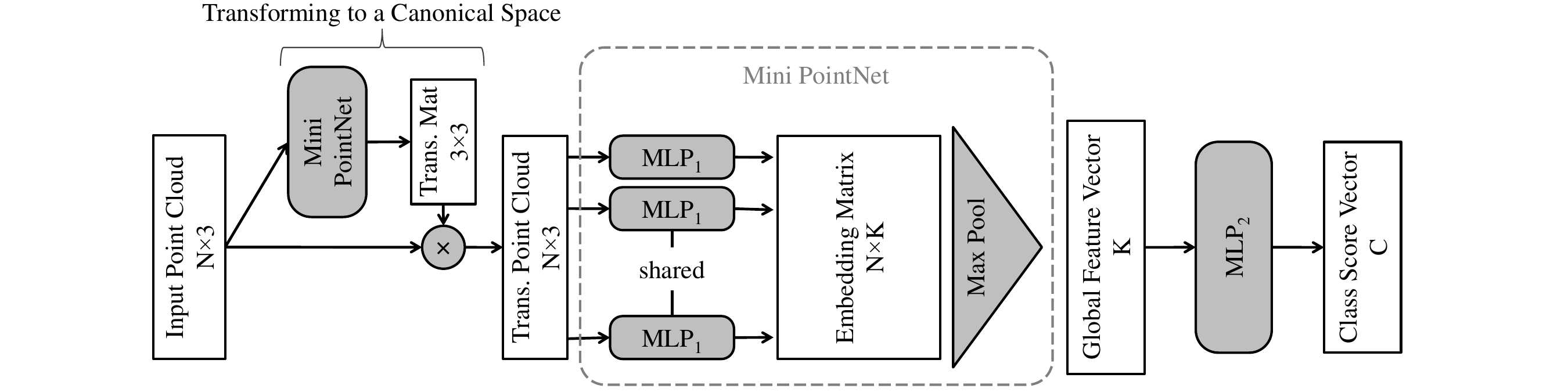}
    \caption{A simplified architecture of PointNet (2017) \cite{Charles17PointNet}. Each of the $N$ points of the input point cloud is projected to a $K>3$ dimensional space and pools the maximum value from each dimension, forming a vector of $K$ features.}
    \label{fig:pointnet-architecture}
    \Description{pointnet-architecture}
\end{figure}

In Figure \ref{fig:pointnet-architecture}, we can see that the points are first transformed to a canonical space with a mini-PointNet (T-Net). Though T-Net does not guarantee true invariance to rotations, translation, scale, and other rigid transformations, it helps to adapt to them to some degree. Then, each of the transformed points is given as input to a shared MLP. MLP encodes the $(x, y, z)$ coordinates to a higher-dimensional embedding space. So, after passing each of the $N$ points through the network, we get a matrix of $N \times K$. Now, a max-pooling layer is used for producing a $K$-dimensional feature vector. This feature vector, called the global feature vector, encodes the relations or interactions of the points with their neighbors. So, this Max pooling layer is mainly responsible for making the network order invariant. Then another MLP is used for the final classification task.

PointNet applies Max Pooling across all points. By doing so, it destroys the shape’s structural information. That is why PointNet cannot capture any local structural feature of the shape. \textbf{PointNet++} \cite{Qi17PointNet++} improves on the network by adding a hierarchical configuration similar to CNN. This hierarchical configuration first extracts features from small local regions, then step-by-step extends to larger regions, and finally yields a global feature that incorporates all local structures.

\begin{figure}[ht]
    \centering
    \includegraphics[width=\textwidth]{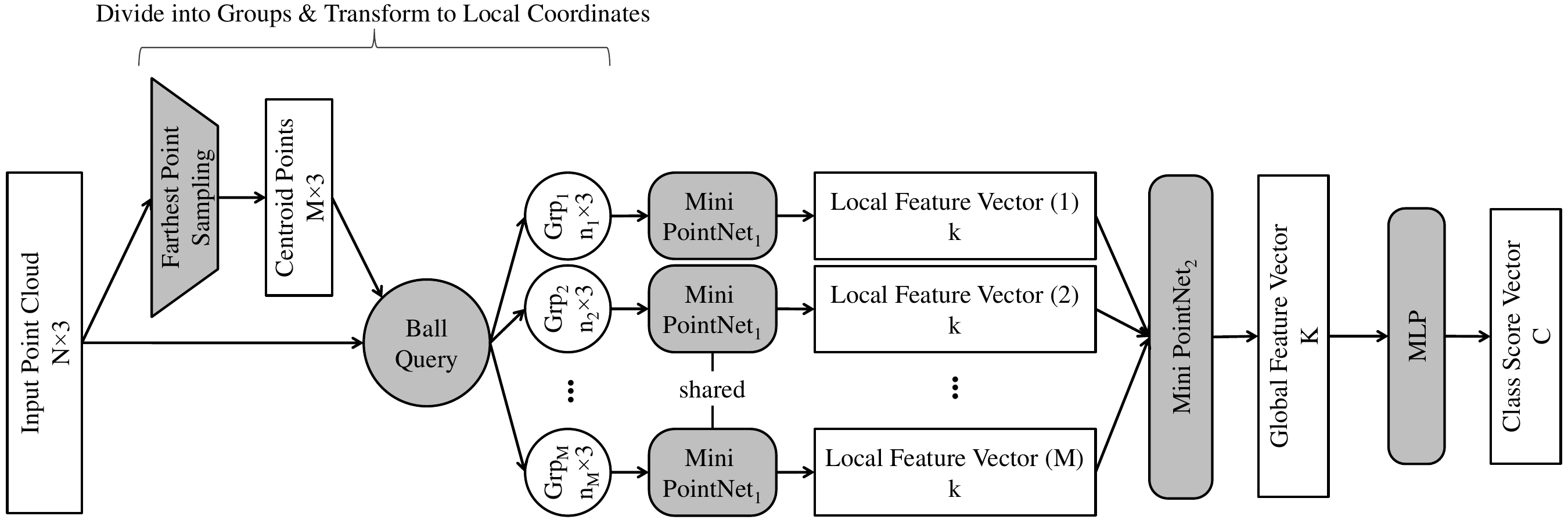}
    \caption{A simplified architecture of PointNet++ (2017) \cite{Qi17PointNet++}. $M$ centroids are sampled from $N$ input points and neighboring points within a fixed radius from the centroids are grouped. A feature vector of size $k$ is extracted from each of the point groups. These local features are later combined for global feature extraction.}
    \label{fig:pointnet++-architecture}
    \Description{pointnet++-architecture}
\end{figure}

PointNet++ uses PointNet as a building block. It starts by applying farthest point sampling \cite{Eldar97FPS} over the pre-normalized point cloud. This produces a subset of points that are called centroids. For each of the centroids, a Ball Query is run, which subdivides the whole point cloud into local point groups. These local point groups’ coordinate system is adjusted to the local reference point by subtracting their coordinates by their respective centroids. Each of these local point groups is then passed through a shared PointNet that extracts local features. In practice, PointNet++ stacks multiple layers of this sampling, grouping, and PointNet for hierarchical local feature extraction. 

Finally, the features are aggregated by another PointNet into a global feature vector, which also encodes the relations among the local structural features of the shape. This is followed by a task head, which can be a fully connected layer for classification tasks or multiple interpolation layers with residual connections (similar to UNet \cite{Ronneberger15UNet}) for segmentation tasks. A visual representation of the PointNet++ is provided in Figure~\ref{fig:pointnet++-architecture}.

PointNet++ requires point cloud normalization, which may not be ideal for real-world data that often includes background noise and irregularities. Additionally, the architecture lacks inherent adaptability to geometric transformations such as rotation, translation, and scaling. Furthermore, its reliance on max-pooling, while effective for capturing dominant features, can lead to the loss of subtle but potentially important information, which may negatively impact downstream tasks. To combat these shortcomings, several subsequent works, such as PointWeb (2019) \cite{Zhao19Pointweb}, PointASNL (2020) \cite{Yan20Pointasnl}, PointMLP (2022) \cite{Ma22PointMLP}, and PointNeXt  (2022) \cite{Qian22PointNext} have proposed improvements over PointNet++ while using PointNet as a building block.

\subsection{Graph Convolution Network}
\label{subsec:graph-convolutional-network}

Graph Convolution Networks (GCNs) \cite{Kipf16Semi, Niepert16Learning} such as RGCNN (2018) \cite{Te18RGCNN}, DGCNN (2019) \cite{Wang19dgcnn}, PointRGCN (2019) \cite{Zarzar19PointRGCN}, GeomGCNN (2021) \cite{Srivastava21GEOMGCNN}, and PointViewGCN (2021) \cite{Mohammadi21PointviewGCN} have been very successful in processing point clouds. GCNs are a class of DL procedures tailored for operating on graph-structured data where the nodes represent entities or objects and the edges represent interactions between them. The key idea behind GCN is the generalization of the convolution operation for both regular and irregular types of data. This enables them to effectively source both local and global relationships present in a graph structure. GCN achieves this by iterative information aggregation. It usually contains alternating layers of message-passing and non-linear activation functions. This allows the network to learn complex representations of the nodes and relationships between them.

\begin{figure}[ht]
    \centering
    \includegraphics[width=0.96\textwidth]{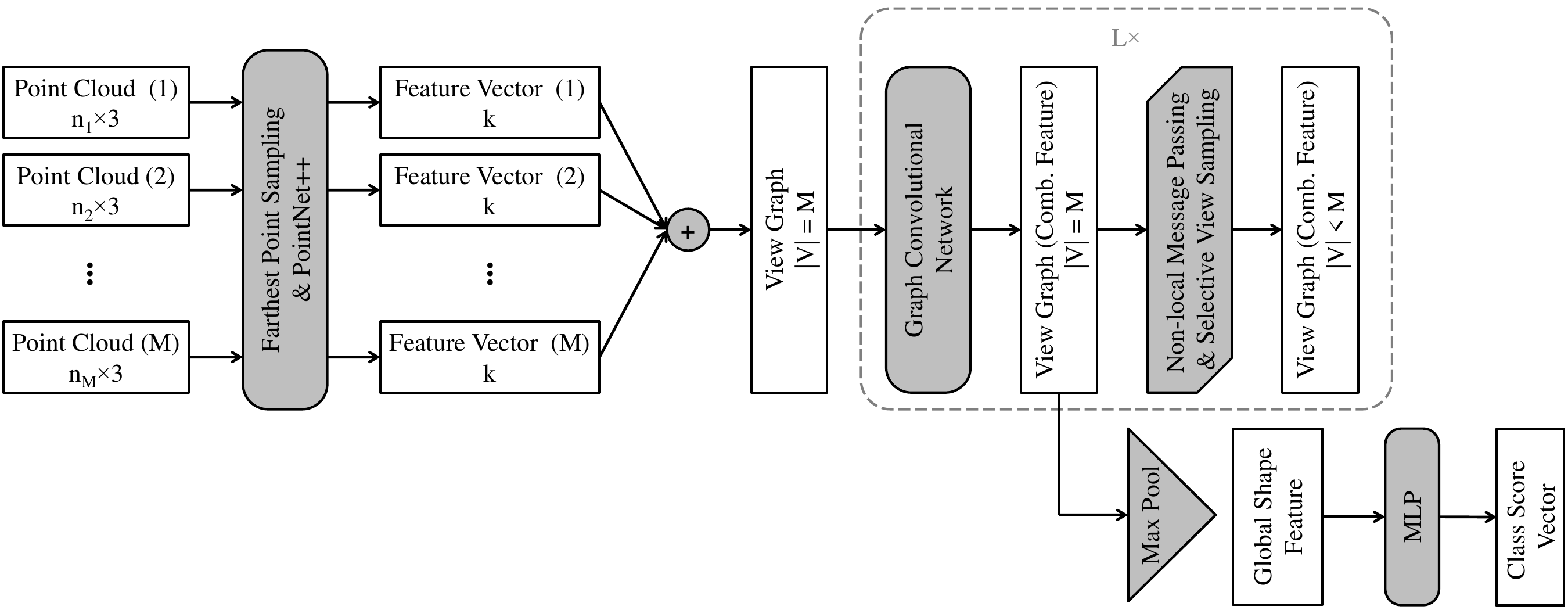}
    \caption{A simplified architecture of PointViewGCN (2021) \cite{Mohammadi21PointviewGCN}. $n$ points are sampled from each of the $M$ scans captured by a depth sensor. A graph with $M$ nodes is formed from the feature vectors (size $k$) of these point clouds, depending on their viewpoints. The graph then goes through $L$ GCN layers for final feature extraction.}
    \label{fig:pointviewgcn-architecture}
    \Description{pointviewgcn-architecture}
\end{figure}


\textbf{PointViewGCN} ingests a set of point cloud data captured from different viewpoints around the object. Each point cloud is put through a farthest point sampling (FPS) \cite{Eldar97FPS} algorithm for sampling the most feature-rich 210 points. Then, a pretrained PointNet++ \cite{Qi17PointNet++} is used for extracting the shape's feature vector of size 512. Then, a view graph is constructed by aggregating the features of the input point clouds, where each node is represented by the feature vector of one single view point cloud. A node is considered connected to another node depending on their view positions. So, if two views are taken side by side by the sensor then the corresponding nodes are connected. 

The resulting graph is then processed by a multi-level GCN. The first level consists of a standard local GCN and a non-local message passing, followed by selective view sampling. The local graph convolution updates the features of each node by infusing features from its neighboring nodes. Non-local message passing updates the features of each node by combining features from all other nodes in the entire graph. Lastly, selective view sampling (SVS) is used for removing less descriptive views/nodes from the graph. SVS first applies FPS to sub-sample nodes from the graph. From each sub-sampled node's neighborhood, one node is selected that produces the highest response (value) after applying softmax. This new sub-sampled graph is then fed to the next level of GCN and SVS. 

The features obtained at the local GCN of each level are passed through Max-Pooling for calculating the global shape feature. These global shape features derived from each level are concatenated and put through a Fully Connected Network for Global Loss calculation. An overview of the above-discussed functions is presented in Figure \ref{fig:pointviewgcn-architecture}.

\new{\textbf{DGCNN} (Dynamic Graph Convolutional Neural Network) \cite{Wang19dgcnn} is another landmark model that integrates ideas from both PointNet and graph-based learning instead of relying on traditional GCN layers. Its core contribution is the EdgeConv module, which captures local geometric relationships by aggregating features from neighboring points and can be stacked to progressively learn higher-level shape information. The overall architecture is straightforward, consisting of multiple EdgeConv layers followed by a global pooling layer that produces a compact representation of the entire point cloud.}

\subsection{Transformer}
\label{subsec:transformer}

Recently, transformer-based architectures \cite{Vaswani17Transformer} have achieved state-of-the-art performance across a wide range of machine learning tasks, setting new benchmarks in multiple domains. Though first employed for language translation, the self-attention mechanism of transformers has been proven very useful in deriving and accumulating relations among the different fragments of an input data. PCT (2021) \cite{Guo21PCT}, PointTransformer (2021) \cite{Zhao21PointTransformer}, PointBERT (2022) \cite{Yu22PointBERT}, ReCon (2023) \cite{Qi23ReCon}, PointGPT (2024) \cite{Chen24PointGPT},  PTV3 (2024) \cite{wu24ptv3}, and ReCon++ (2024) \cite{Qi24ShapeLLM} are some of the well-recognized transformer architectures employed for point cloud processing.

\begin{figure}[ht]
    \centering
    \includegraphics[width=0.96\textwidth]{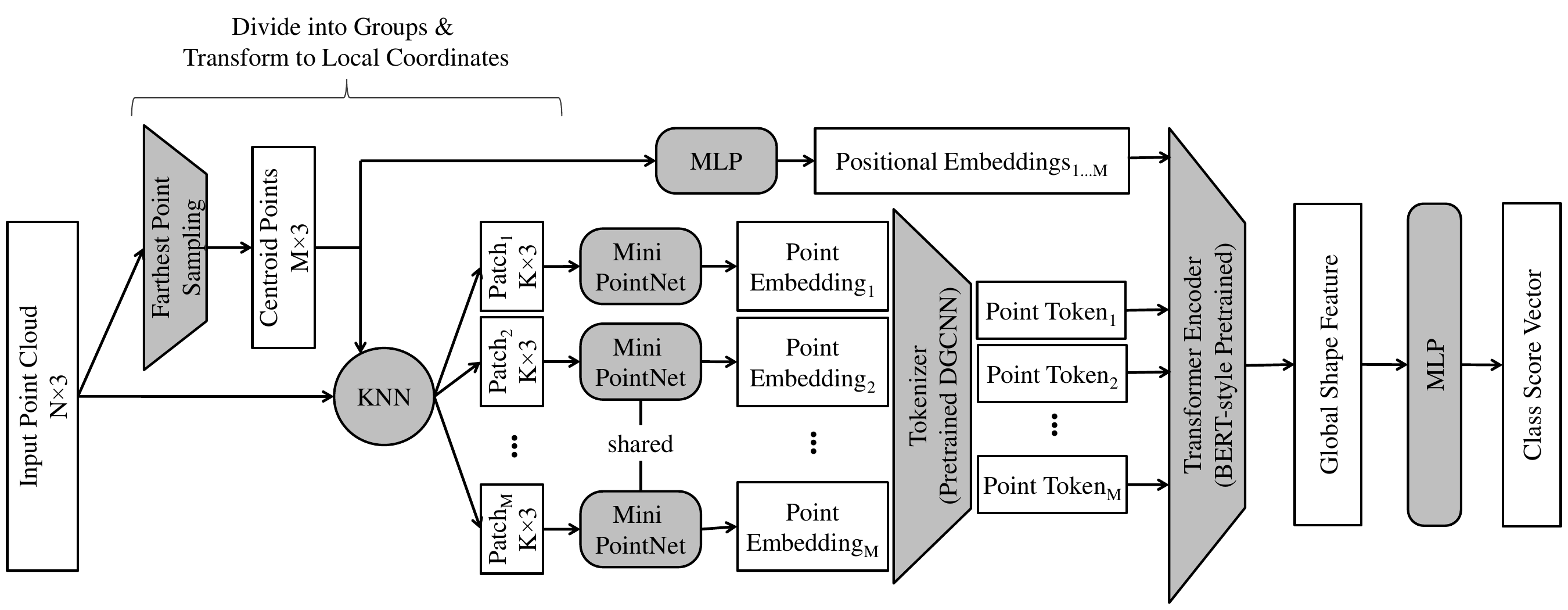}
    \caption{A simplified architecture of PointBERT (2022) \cite{Yu22PointBERT}. $M$ points are selected from a point cloud of $N$ points, each used for extracting a local point patch of $K$ points. The patches are then processed by an embedding layer and a tokenizer. Finally, a BERT-style transformer extracts features from the tokens.}
    \label{fig:pointbert-architecture}
    \Description{pointbert-architecture}
\end{figure}

Figure \ref{fig:pointbert-architecture} presents a simplified structural design of \textbf{PointBERT}. From the name, we can understand that it employs a BERT-style \cite{Devlin18BERT} pretrained transformer for encoding point sets. First, the researchers sample some points (centroids) from the point cloud using farthest point sampling \cite{Eldar97FPS}. Then, k-nearest neighbor (KNN) is used to create groups of points for each centroid. These point groups are called local point patches, and they encode the local structure of the shape. The coordinates of the points of each path are subtracted by their corresponding centroids. This shifts the point patches to the local reference point and so makes them unbiased by the overall structure.

Next, these point patches are fed through a mini-PointNet \cite{Charles17PointNet} that projects the point patches to point embeddings (pretty similar to the PointNet++). These point embeddings are then fed into a discrete variational autoencoder (discrete VAE) \cite{Kingma13VAE, Rolfe16discrete}, which is tasked with respective point patch reconstruction. DGCNN \cite{Wang19dgcnn} is employed as the encoder of the discrete VAE. Once training is complete, the latent space representations produced by the encoder serve as compact tokens for the respective point patches. As a result, the decoder is no longer required during inference, and the encoder effectively functions as a tokenizer.

The positional embedding of each point patch is obtained by applying an MLP on each respective centroid’s global coordinate. This is because the point patch’s position in the global space is represented by the respective centroid’s position. MLP transforms the coordinates of the centroids into a higher-dimensional space. This transformation effectively creates a unique positional embedding for each point patch.

Now, all the point tokens, positional embeddings, and the class token are sent directly to a BERT-style \cite{Devlin18BERT} transformer for pretraining. The input point tokens are randomly replaced with predefined and learnable mask tokens and then fed to the transformer. The transformer learns to recover the missing point tokens through contrastive learning, enabling it to infer the geometric structure of the shape. There can be multiple layers of transformer blocks, and the last layer outputs the global feature.

\new{\textbf{PTv3} (Point Transformer V3) \cite{wu24ptv3} is a highly efficient and scalable point-cloud processing backbone that builds upon the strong foundations laid by PTv1 \cite{zhao21ptv1} and PTv2 \cite{wu22ptv2}. It is built on the philosophy of prioritizing simplicity and efficiency over the accuracy of less impactful computation. It replaces costly KNN graph construction with serialized neighborhood mapping using space-filling curves, enabling structured processing of unstructured point clouds. It then performs window-based attention on contiguous point patches and uses lightweight conditional positional encoding (xCPE) via sparse convolutions to capture geometric context. Additionally, PTv3 introduces lightweight patch attention and enhanced conditional positional encoding (xCPE), eliminating costly relative positional encodings. This dramatically expands the receptive field while reducing memory and computation.}

\begin{figure}[ht]
    \centering
    \includegraphics[width=0.93\textwidth]{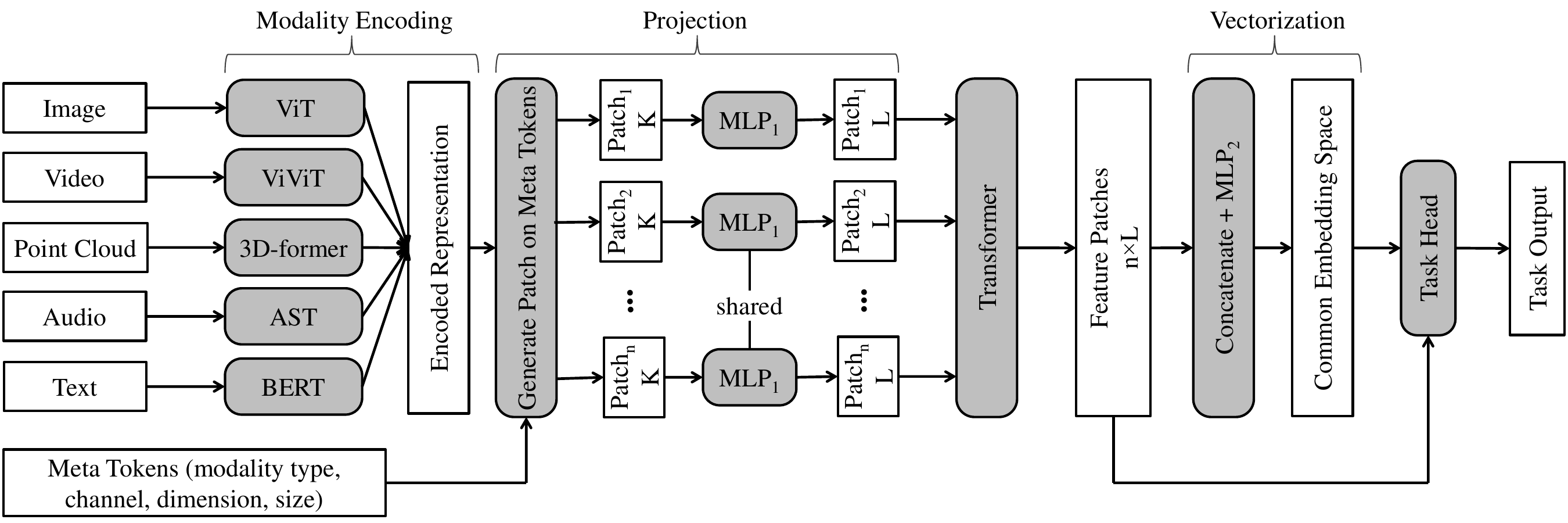}
    \caption{A simplified architecture of OmniVec (2024) \cite{Srivastava24OmniVec}. A separate pretrained transformer model is used to encode each modality. Depending on the task and the modality, the encoding is divided into $n$ patches. Each patch (size $K$) goes through a projection layer and a transformer layer. Then, the patches are concatenated and used as features for downstream tasks.}
    \label{fig:omnivec-architecture}
    \Description{omnivec-architecture}
\end{figure}

Most of the previously discussed methods – and in fact majority of machine learning approaches – are designed to be task and modality specific. This specialization fragments the available data across isolated domains, limiting the ability of models to generalize. As a result, tasks and domains with limited data are disproportionately affected, often leading to sub-optimal performance. Additionally, learning various tasks together in multiple modalities in a unified environment can create a unique regularization effect that is not possible to achieve in a single-task single-modality setting. As a large number of parameters are shared while learning different tasks in different modalities, it is more likely that the network will learn a more meaningful representation from data without over-fitting to one specific task or modality. Learning from multiple modalities for multiple tasks also helps in leveraging available data from multiple domains in compensating for the unavailability of labeled large-scale datasets from other domains. These requirements suit very well for point cloud processing tasks. We can see in Figure \ref{fig:model-comparison} that \textbf{OmniVec} \cite{Srivastava24OmniVec} dominates 3D vision tasks over other methods by a significant margin. OmniVec introduces modality specific encoder and a shared network that can ingest the encodings from different modalities and learn a more abstract representation from both spatial and temporal training data. This allows sharing of knowledge learned from different datasets and thus improving performance in all downstream tasks (see Figure \ref{fig:omnivec-architecture}).

\subsection{Others Notable Architectures}
\label{subsec:others-notable-architectures}

Several other notable methods have introduced novel architectures demonstrating competitive performance in point cloud understanding tasks. For instance, \textbf{RCNet} \cite{Wu19RCNet} presents a permutation invariant method to capture spatial structure by subdividing the point clouds and then sorting them. Each subdivision of point cloud is then processed by a Recurrent Neural Network (RNN) \cite{Elman90finding, Hochreiter97LSTM}. The resulting features are fed into a regular 2D CNN for global feature extraction. Another RNN-based model is \textbf{Point2Sequence} \cite{Liu19Point2Sequence}. It first extracts local features from the point cloud at multiple scales. These multi-scale features are treated as sequences that capture the hierarchical structure of the whole point cloud, which are then processed by an RNN-based encoder-decoder.

\textbf{Kd-Net} (2017) \cite{Klokov17KdNet} constructs multiple k-dimensional trees from a point cloud with different data splitting directions. Through this, the model hierarchically organizes the points in 3D space. The network starts by learning features from its leaf nodes and, step by step, aggregating them back to the root node. This process allows the model to extract the hierarchical structure of the input point cloud. Thus, the feature representation accumulated at the root node encapsulates the entire point cloud’s shape information. This global feature vector is then used for down-stream tasks. 

\textbf{PVNet} (2018) \cite{You18PVNet} learn from both 3D point cloud and 2D projections. The researchers first extract a global feature vector from the multi-view projections of a 3D shape and then project the feature vector back onto the point cloud space. This projected feature vector is then fused with the point cloud features which is utilized for shape understanding tasks.

\textbf{ULIP} (2023) \cite{Xue23ULIP} recommends a 3D data learning framework which addresses the challenge of limited 3D data size. It leverages a pretrained vision-language model, which is able to encode a 2D image and text in the same representation space. ULIP learns to also encode 3D image in that representation space and thus boost the accuracy of classification tasks. Thus, feature representations learned from large-scale 2D datasets are leveraged in 3D vision tasks. This paradigm of joint understanding is later more exploited by \textbf{OmniVec} \cite{Srivastava24OmniVec} and \textbf{DINO} \cite{abou25dino}.

\textbf{Sonata} (2025) \cite{wu25sonata} has set new SOTA results across various 3D indoor and outdoor perception tasks by introducing a robust self-supervised learning framework. They employ self-distillation at coarser spatial scales and progressively increase the task difficulty. Their approach enables rich feature learning by overcoming the "geometric shortcut" problem, where models rely on low-level spatial cues instead of meaningful representations.

Recently, state-space models such as Mamba \cite{gu2024mamba} have gained traction in 3D data analysis due to their ability to model long-range dependencies with linear computational complexity. The authors of \textbf{Pamba} (2024) \cite{li2025pamba} adapt Mamba to unordered point clouds through a multi-path serialization strategy and a ConvMamba block that enhances local geometry modeling and enables bidirectional information flow. It processes entire point clouds without patching, demonstrating superior scalability, accuracy, and generalization to transformer-based models across several indoor and outdoor benchmarks. \textbf{Serialized Point Mamba} (2024) \cite{wang2024serialized} improves on the latency as well by enhancing efficiency through bidirectional modeling, conditional positional encoding, and grid pooling.

\section{Evaluation of Point Cloud Processing Architectures}
\label{sec:evaluation-of-point-cloud-processing-arch}

In this study, we adopt \textbf{overall accuracy} (OA) for classification and \textbf{mean Intersection over Union} (mIoU) for segmentation, as these are the most widely used evaluation metrics in the literature. A meaningful assessment, however, should also account for qualitative properties such as model robustness, scalability, and generalization capability; additionally, dataset-specific factors, such as class imbalance, sample distribution variability, and the extent to which the dataset reflects real-world conditions and deployment scenarios.

\subsection{Evaluation Metrics for Classification}
\label{subsec:evaluation-metrics-for-classification}
For comparing the performance of different classification models, researchers use various numerical scores such as accuracy, precision, and recall to reveal particular complementary insights into a model’s strengths and weaknesses. Given special circumstances, like- class imbalance or high cost for false predictions, more complex metrics, including F1 score and Area Under the Receiver Operating Characteristic Curve (AUC-ROC), are used to provide robust assessment. However, the most common performance measure used for comparing models for 3D shape classification is overall accuracy (OA) and mean class accuracy (mAcc) \cite{Guo21Deep-dl3dsurvey}. OA is the ratio of the total number of correct predictions (True Positive ($TP$) + True Negative ($TN$)) and the total number of predictions (True Positive ($TP$) + False Positive ($FP$) + True Negative ($TN$) + False Negative ($FN$)) (see equation \ref{eq:OA}).

\begin{equation}
OA = \frac{TP+TN}{TP+FP+TN+FN} \times 100\% 
\label{eq:OA}
\end{equation}

Many 3D benchmark datasets have class imbalance, containing a disproportionately large number of samples from dominant object categories such as chair and table, while underrepresented classes such as bathtub and TV have substantially fewer instances. Consequently, OA can be misleadingly high, even when a model performs poorly on minority classes.  mAcc is calculated by first calculating the accuracy for each shape class ($i$) and then averaging them (see equation \ref{eq:mAcc}). Therefore, mAcc offers a better understanding of how the model is able to deal with classes having uneven sample sizes.

\begin{equation}
mAcc = \frac{1}{N} \sum_{i=1}^{N} \left(\frac{TP_i}{TP_i+FN_i}\right) \times 100\% 
\label{eq:mAcc}
\end{equation}

\subsection{Evaluation Metrics for Segmentation}
\label{subsec:evaluation-metrics-for-segmentation}
Classwise OA and mAcc are also used for appraising models on 3D scene segmentation. However, the most widespread metric is- mean intersection over union (mIoU). mIoU calculates the overlap between the predicted region ($Pred$) and the ground truth ($GT$) region for each class $i$ (see equation \ref{eq:mIoU}).

\begin{equation}
mIoU = \frac{1}{N} \sum_{i=1}^{N} \left(\frac{| Pred_i \cap GT_i |}{| Pred_i \cup GT_i |}\right) \times 100\%
\label{eq:mIoU}
\end{equation}

\subsection{Model Performance Assessment}
\label{subsec:model-performance-assessment}

As shown in Table \ref{tab:model-comparison}, SOTA models achieve over 96\% overall accuracy for point cloud classification, which appears promising at first glance. However, this performance must be interpreted with caution. Benchmark datasets such as ModelNet40 and ScanObjectNN include only 40 and 15 object categories, respectively, and suffer from significant class imbalance. Consequently, while overall accuracy remains high, the mean class accuracy is notably lower, with rare or small object categories often exhibiting classification accuracies below 50\%. This disparity indicates that models may overfit to dominant classes and fail to generalize across the full label space.

\begin{table}
\caption{Comparison among network models for point cloud classification and segmentation tasks. Data in bold represents the best performance; the second-best is underlined.}
\label{tab:model-comparison}
\resizebox{\textwidth}{!}{
\begin{tabular}{ccccccc}
  \toprule
  \textbf{\makecell{Network \\ Architectures}}
  & \textbf{Year ↑}
  & \textbf{Architecture Type}
  & \textbf{\makecell{Classification \\ ModelNet40 \\ (OA \%)}}
  & \textbf{\makecell{Classification \\ ScanObjectNN \\ (OA \%)}}
  & \textbf{\makecell{Part Seg. \\ ShapeNetPart \\ (mIoU \%)}}
  & \textbf{\makecell{Semantic Seg. \\ S3DIS \\ (mIoU \%)}}\\
  \midrule
  3DShapeNets \cite{Wu153DShapeNets} & 2015 & Convolutional Network & 77.0 & - & - & - \\
  MVCNN \cite{Su15MVCNN} & 2015 & Convolutional Network & 90.1 & - & - & - \\
  PointNet \cite{Charles17PointNet} & 2017 & Multi-Layer Perceptron & 89.2 & 68.2 & 83.7 & 47.6 \\
  PointNet++ \cite{Qi17PointNet++} & 2017 & Multi-Layer Perceptron & 90.7 & 77.9 & 85.1 & 54.5 \\
  PointCNN \cite{Li18PointCNN} & 2018 & Convolutional Network & 92.2 & 78.5 & 86.1 & 65.4 \\
  DGCNN \cite{Wang19dgcnn} & 2019 & Graph Network & 93.5 & 78.1 & 85.2 & 56.1 \\
  PointConv \cite{Wu19PointConv} & 2020 & Convolutional Network & 92.5 & - & 85.7 & - \\
  GeomGCNN \cite{Srivastava21GEOMGCNN} & 2021 & Graph Network & \underline{95.9} & - & \textbf{89.1} & 70.1 \\
  PointTransformer \cite{Zhao21PointTransformer} & 2021 & Transformer & 93.7 & - & 86.6 & 73.5 \\
  PointViewGCN \cite{Mohammadi21PointviewGCN} & 2021 & Graph Network & 95.4 & 85.5 & - & - \\
  PointBERT \cite{Yu22PointBERT} & 2022 & Transformer & 93.8 & 83.1 & 85.6 & - \\
  PointMLP \cite{Ma22PointMLP} & 2022 & Multi-Layer Perceptron & 94.5 & 85.7 & 86.1 & - \\
  PointNext \cite{Qian22PointNext} & 2022 & Multi-Layer Perceptron & 94.0 & 87.7 & \underline{87.1} & 74.9 \\
  ReCon \cite{Qi23ReCon} & 2023 & Transformer & 94.7 & 91.3 & 86.4 & - \\
  OmniVec \cite{Srivastava24OmniVec} & 2023 & Transformer & \textbf{96.6} & \textbf{96.1} & - & \underline{75.9} \\
  ReCon++ \cite{Qi24ShapeLLM} & 2024 & Transformer & 95.0 & \underline{95.3} & - & - \\
  Sonata + PTv3 \cite{wu25sonata, wu24ptv3} & 2025 & Transformer & - & - & - & \textbf{82.3} \\
  \bottomrule
\end{tabular}
}
\end{table}

Furthermore, unlike 2D images, where objects are often entangled with complex backgrounds and exhibit significant appearance variation across different viewpoints, 3D point cloud data inherently captures the explicit spatial representation of objects, independent of lighting, texture, and background clutter, offering a more consistent basis for shape understanding. However, when compared to SOTA performance in 2D image classification, 3D models still lag behind in both accuracy and robustness.

A similar gap exists in the domain of segmentation. While 2D semantic segmentation has reached a high level of maturity, with models demonstrating strong generalization and real-world applicability, 3D segmentation methods continue to face challenges in handling sparse data, occlusions, and varying point densities. Bridging this gap requires fundamental advancements in 3D representation learning, architectural design, and data augmentation strategies.

\section{Future Research Directions}
\label{sec:future-research-directions}

As outlined in Section~\ref{subsec:model-performance-assessment}, significant challenges remain in 3D shape processing, particularly in achieving the robustness, efficiency, and generalization seen in 2D vision systems. Below, we highlight several emerging and promising research directions that aim to address these limitations:

\subsection{Self-Supervised Learning}
\label{subsec:self-supervised-learning}
One of the primary reasons 3D image processing continues to lag behind its 2D counterpart is the limited availability of large-scale, high-quality annotated datasets. While 2D vision has benefited immensely from extensive benchmarks such as ImageNet\cite{Deng09Imagenet} and COCO \cite{lin2014microsoftcoco}, the 3D domain lacks datasets of comparable scale, diversity, and annotation richness. As data procurement is expensive, leveraging unlabeled 3D data through pretraining, such as representation learning, is a promising direction \cite{Sauder19SelfSupervised, Zhang21SelfSupervised}. In this paradigm, models are first trained to capture semantically meaningful features, and later fine-tuned on downstream tasks with limited labeled data. This approach not only enhances generalization but also alleviates the challenges posed by the scarcity of large-scale annotated 3D datasets.

\subsection{Efficient Attention Mechanism}
\label{subsec:efficient-attention-mechanism}
Effectively handling outliers, artifacts, and variations in point density remains a significant hurdle for real-world applicability. Transformers have shown great promise in extracting high-level point cloud features while handling real-world challenges, but they prove computation heavy especially for large-scale point clouds. Developing more sophisticated attention mechanisms specifically tailored to the unique characteristics of point clouds could lead to improved performance. Future work will likely focus on sparse attention mechanisms, point-cloud-specific tokenizations, and quantized architectures to reduce memory or compute while maintaining performance.

\subsection{Scalable and Generalizable Network Architectures}
\label{subsec:scalable-network-architectures}
Despite the substantial advancements in GPU capabilities and modern hardware acceleration, 3D data processing remains computationally intensive. As a result, a majority of existing point cloud processing pipelines incorporate a sampling step to reduce computational overhead by selecting a smaller subset of points from the original data. This inevitably results in the loss of potentially valuable geometric and semantic information captured by the sensor. Developing more efficient and scalable architectures capable of directly processing large-scale, high-density point clouds can lead to significant improvements in the robustness, versatility, and generalization ability of models across different datasets and application domains.

\subsection{Context Aware Learning}
\label{subsec:graph-neural-networks}
Real-world 3D scenes exhibit complex interdependencies among their constituent elements, such as objects, architectural structures, and supporting geometry. Graph neural networks (GNNs), and in particular Graph Convolutional Networks (GCNs), have demonstrated an ability to model these high-level relationships, yielding performance gains in various 3D applications. However, their potential remains underexplored in the domains of semantic segmentation and object classification. A systematic investigation of GCN-based approaches for encoding the intricate relational patterns among point sets may therefore unlock substantial improvements in these tasks.

\subsection{Multi-Modal and Multi-Sensor Fusion}
\label{subsec:multi-modal-and-multi-sensor-fusion}
Modalities such as 2D images, audio, text, and video are significantly more abundant and accessible than 3D data, primarily due to the widespread proliferation of consumer devices, such as smartphones, webcams, and microphones, and the ubiquity of internet platforms, including social media, digital publishing, and streaming services. In contrast, 3D datasets remain limited due to their reliance on specialized hardware, which is not yet widely integrated into mainstream consumer workflows. This scarcity presents a significant bottleneck in training DL models that generalize well across tasks and domains, ultimately hindering the development of robust and transferable 3D representations. Therefore, exploring the combination of point cloud with other modalities could lead to a more comprehensive understanding of 3D scenes. Networks that jointly process 2D and 3D inputs using cross-attention are a growing trend. Similarly, fusing LiDAR and radar or combining multiple LiDAR scans are open problems. Another direction is Cross-Modal sharing- sharing weights between modalities for leveraging knowledge learned from a different modality for a different task.


\subsection{Interaction and Motion Estimation}
\label{subsec:interaction-and-motion-estimation}
Understanding and predicting the interaction dynamics and motion trajectories of 3D objects is the next reasonable step in comprehensive 3D scene understanding. This involves not only estimating how an object moves in space but also modeling state transitions, for instance, opening a drawer, closing the lid of a laptop, or sliding a window. Also, the manipulation of deformable or non-rigid objects, such as \new{human body,} mattresses, plush toys, or fabric materials, is another area for future research. Additionally, recent research has begun to explore material and texture inference from 3D data, which facilitates higher-level scene understanding tasks such as acoustic prediction. For example, the ability to simulate the sound generated by two wooden surfaces sliding against each other.


\subsection{Other Research Directions}
\label{subsec:other-research-directions}

There are significant gaps in the standardization of point cloud data compression and encoding, which hinder interoperability and efficient storage. Furthermore, there is a pressing need for more advanced sensor technologies capable of capturing high-fidelity, real-world 3D data in diverse environments. The development of large-scale datasets with rich annotations and broad category coverage is equally critical to enable the training of robust and generalizable models. 


\new{To address data scarcity and annotation costs, recent research in point cloud segmentation has increasingly turned toward few-shot learning paradigms \cite{zhao21fewshot, fayjie25fewshot}. Another emerging and impactful line of work on point cloud semantic segmentation is hierarchical learning; understanding that a "chair" is also a "furniture" item and has "legs" \cite{Li25DeepHierarchical}. This introduces a probabilistic representation of class hierarchies and regularization to model the intrinsic multi-granularity of 3D scenes.} Additionally, the promising capabilities of recent state-space models \cite{Gu23Mamba} need to be further explored for point cloud analysis. 

\new{Advancing end-to-end frameworks for 3D object detection and tracking is similarly crucial, especially as current models continue to struggle with small objects, transparent materials, and reflective surfaces.} Finally, point cloud generation remains a nascent area and requires significant advancements to achieve practical utility \cite{Luo21DPM, Zeng22LION, nichol22point-e}.

\section{Conclusion}
\label{sec:conclusion}

This article will serve as a valuable resource for both learners and practitioners starting to work in 3D vision. Here, we have investigated the unique difficulties posed by the unstructured and unordered nature of point cloud data. And then, we have provided a taxonomy of the most influential and cutting-edge DL approaches for 3D point cloud classification and segmentation. For each category, we have presented a synopsis of the notable works and have gone into a deep discussion about at least one prominent solution from every category with an easy-to-understand diagram and a simplified architectural description. There is also a performance comparison among the solutions on widely recognized benchmarks, which provides a valuable insight into their applicability.

Despite substantial progress, most existing methods deal only with small point sets, often sampled from the original data. However, modern sensors yield dense and comprehensive point clouds, sometimes an entire model of a city block. This highlights the need for more scalable and efficient algorithms for 3D data processing. We also lack large-scale real-world datasets for point clouds with rich annotations that cover a wide range of object categories. And, we still require standardized practices in point cloud encoding formats.

\subsubsection*{Acknowledgments}
This work was supported by NSF \#2532731 and \#2603236. Opinions, findings, and conclusions expressed herein are those of the authors and do not necessarily reflect the views of the NSF.

\bibliographystyle{ACM-Reference-Format}
\bibliography{references}

\newpage

\renewcommand{\thesubsubsection}{\thesection.\arabic{subsubsection}}
\appendix
\section{Other 3D Structure Representation Modalities}
\label{sec:other-3d-structure-representation-modalities}

Regular 2D (raster) images are represented by a matrix of pixels, which is by far the most prevalent among the other 2D image representation methods. There might be different compression and encoding formats like- JPEG, PNG, BMP, etc. But the core representation method is the same- a matrix of pixels. However, a precise yet efficient representation of 3D scenes remains an unsolved problem. Figure~\ref{fig:3d-representation} illustrates the commonly used 3D geometric data representation formats, each offering distinct advantages and limitations, making them suitable for specific tasks and application domains:


\subsubsection{Depth Map}
\label{subsubsec:depth-map}
Encodes the distance of each pixel in a 2D image from the camera. While visually similar to regular 2D images, depth maps differ in that their pixel values represent distances rather than colors or intensity. 
Because depth maps are structured as a 2D grid and do not inherently represent complete three-dimensional structure, they are often characterized as 2.5D representations. Most sensors used for 3D scanning produce depth maps, which are later combined together for constructing more complex 3D representations.


\subsubsection{Mesh}
\label{subsubsec:mesh}
Uses point set (analogous to point cloud) connected via polygons, typically triangles, for describing the 3D surface of objects. Synthetic meshes are produced by human graphic designers with the help of CAD software. There also exist various algorithms, as well as ML approaches, that are employed to construct meshes from point clouds. Meshes are used heavily in computer graphics and modeling, but are nontrivial to process with ML architectures.

\subsubsection{Voxel Grid}
\label{subsubsec:voxel-grid}
Abbreviated from volumetric-pixel, it models the whole shape as a 3D grid of discrete occupancies. While being simple and effective for capturing volumetric data, voxel grids cause high memory costs in depicting fine details. And with higher resolution, the memory demand increases exponentially. Octree-based representation, a variation of voxel grid that exploits hierarchical structure to create voxels of different sizes, can capture details while being memory efficient. However, this comes at the cost of increased data complexity and higher computational overhead. Both voxel grids and octrees are primarily derived from enclosed 3D meshes.

\subsubsection{Surface Normals}
\label{subsubsec:surface-normals}
Provides a directional vector at each 3D coordinate, indicating the local surface orientation in the XYZ plane. Normals are not directly captured by sensors, but can be computed from depth maps, point clouds, or mesh data. They are often pre-calculated for a variety of applications like- photorealistic rendering (e.g., lighting and shading), physics-based simulation, and collision detection in animation.

\subsubsection{Others}
\label{subsubsec:others}
To better address the diverse requirements of various tasks and domains, several alternative 3D representation modalities have been explored beyond conventional formats. These include 2D projections, which simplify 3D processing by flattening spatial data into planar representations, such as Mercator projections (particularly used in facial geometry); multi-view representations, where multiple rendered 2D images from different viewpoints are stored, can be processed using standard 2D CNNs; neural radiance fields (NeRF) \cite{Mildenhall21NeRF}, represents scenes through weights of a neural network as continuous volumetric functions learned from multi-view images, used for photorealistic novel view synthesis; and Gaussian splatting \cite{kerbl233DGS}, which models scenes as a collection of 3D Gaussians (ellipsoids) with color parameters and opacity, offering a balance between data fidelity and rendering efficiency.

\section{Point Cloud Acquisition}
\label{sec:acquisition}

\begin{figure}[ht]
    \centering
    \includegraphics[width=\textwidth]{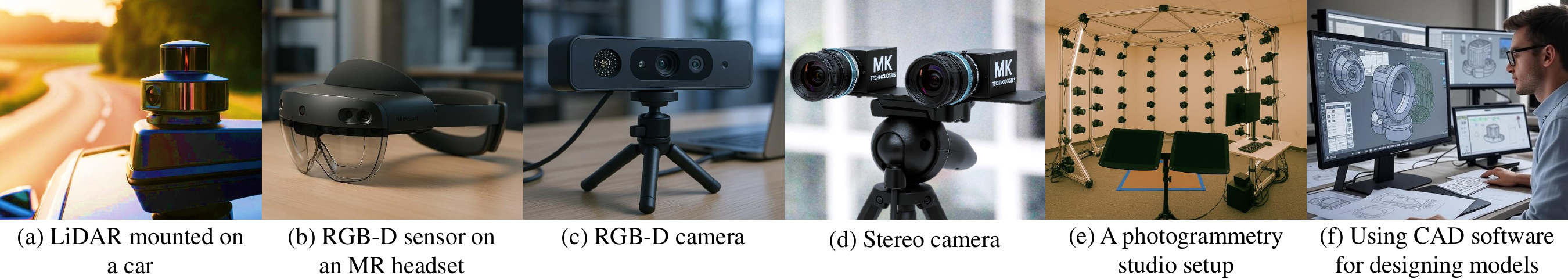}
    \caption{\new{Sources for point cloud data acquisition.}}
    \label{fig:pointcloud-acquisition}
    \Description{pointcloud-acquisition}
\end{figure}

Point cloud is the most common format of raw data coming out of 3D imaging sensors and can also be obtained from other 3D representation modalities. Figure \ref{fig:pointcloud-acquisition} presents the different sensors and software techniques that act as a source for point clouds.

\subsubsection{LiDAR Sensor}
\label{subsubsec:lidar}
Employs an active remote sensing scheme which emits short pulses of laser light onto the scene. Early versions of LiDAR measured the distance of objects by using the time that each pulse took to reflect back from the surface of the objects (time of flight). Current LiDARs use phase-shift detection for achieving the same feat while being more precise and faster. LiDAR sensors are usually used for scanning outdoor environments, and with the increase of distance, the point cloud becomes sparser. There are different types of scanning technologies used under the LiDAR umbrella: Mobile Laser Scanners (MLS)- urban mapping and infrastructure surveys, Aerial Laser Scanners (ALS)- mapping and creating elevation models of large areas, and static Terrestrial Laser Scanners (TLS)- detailed scans of buildings and cultural heritage sites. Adverse weather conditions, such as rain and dust, can significantly affect the performance of LiDAR sensors.

\subsubsection{RGB-D Camera}
\label{subsubsec:rgb-d}
Used for close-range indoor 3D data capture. It is essentially a 2-sensor system- a regular RGB camera and a depth sensor. For depth measurement, it either uses structured light, time of flight (ToF), or a combination of both methods. So, the depth sensor is also an active imaging system that usually emits a pattern of infrared light on objects. The resolution of the depth sensor is usually less than that of the RGB camera. Though point cloud is not the direct result of the RGB-D sensor, we can easily derive a dense point cloud from it. Typically, this sensor has a limited range and does not work on shiny or transparent surfaces, but it is more affordable than LiDAR.

\subsubsection{Stereo Camera}
\label{subsubsec:stereo-camera}
\new{An arrangement of two spatially calibrated cameras to capture synchronized images of the same scene from slightly different viewpoints. The system estimates distance of each pixel by computing the disparity of the corresponding pixels of two images through a process called triangulation. Some stereo setups incorporate an auxiliary infrared (IR) emitter for improving pixel correspondence matching, especially for low-texture or uniform surfaces.}

\subsubsection{Photogrammetry and Videogrammetry}
\label{subsubsec:photogrammetry-and-videogrammetry}
Image processing techniques that estimate the relative distance of target points by comparing overlapping images captured from different angles. It reconstructs the 3D geometry of the scene by applying the same triangulation process used in stereo cameras. Just like RGBD sensors, we can then directly generate a dense 3D point cloud from this. These techniques can be used for both indoor and outdoor environments.

\subsubsection{CAD Models}
\label{subsubsec:cad}
Mesh-based CAD models have a very widespread usage, especially in the context of computer graphics and 3D modeling. Point clouds can be simply sampled from the surface of the CAD models. During the early days of 3D shape processing, when 3D image acquisition sensors were not as readily available, CAD models were the prime sources for point clouds.

\section{Overview of Benchmark Datasets for Point Cloud Processing}
\label{sec:datasets}

This section provides a brief introduction to key publicly available point cloud datasets. Each dataset is characterized by its modality, scale, annotation richness, and domain specificity:


\subsubsection{KITTI (2012) \cite{Geiger12KITTI, Geiger13KITTI}}
\label{subsubsec:kitti}
An urban scene dataset featuring 6 hours of traffic scenarios recorded with a suite of synchronized sensors, including high-resolution stereo cameras and 3D LiDAR scanners mounted on a moving vehicle. The dataset is widely adopted in autonomous driving research as it contains real-world, outdoor, and dynamic environments. It provides rich multimodal annotation for 15k objects from 8 classes: car, van, truck, pedestrian, person sitting, cyclist, tram, and misc., supporting tasks such as object detection, tracking, and pose estimation, monocular depth estimation, scene segmentation, point cloud registration, optical flow estimation, and super-resolution. There are several datasets similar to KITTI such as SemanticKITTI (2019) \cite{Behley19SemanticKITTI}, KITTI-MOTS (2019) \cite{Voigtlaender19KITTIMOTS, Luiten20KITTIMOTS}, and KITTI-360 (2022) \cite{Liao22KITTI360} for different benchmarks.

\subsubsection{NYU Depth V2 (2012) \cite{Silberman12NYUDepthV2}}
\label{subsubsec:nyudepthv2}
Indoor video sequences were captured through Microsoft Kinect across 464 distinct indoor environments from 3 U.S. cities. It includes 1.5k annotated pairs of aligned RGB and depth images where each pixel is labeled with both semantic and instance-level information, and an additional 400k unlabeled frames.

\subsubsection{Sydney Urban Objects (2013) \cite{Deuge13SydneyUrbanObjects}}
\label{subsubsec:sydneyurbanobjects}
Designed for 3D object detection in outdoor urban environments and ideal for 3D scene understanding in autonomous driving scenarios. Data is captured by a mobile LiDER and categorized under 26 classes from 631 objects, which is more exhaustive than KITTI. However, only the popular 14 classes containing 588 objects are used for testing.

\begin{figure}[ht]
    \centering
    \includegraphics[width=\textwidth]{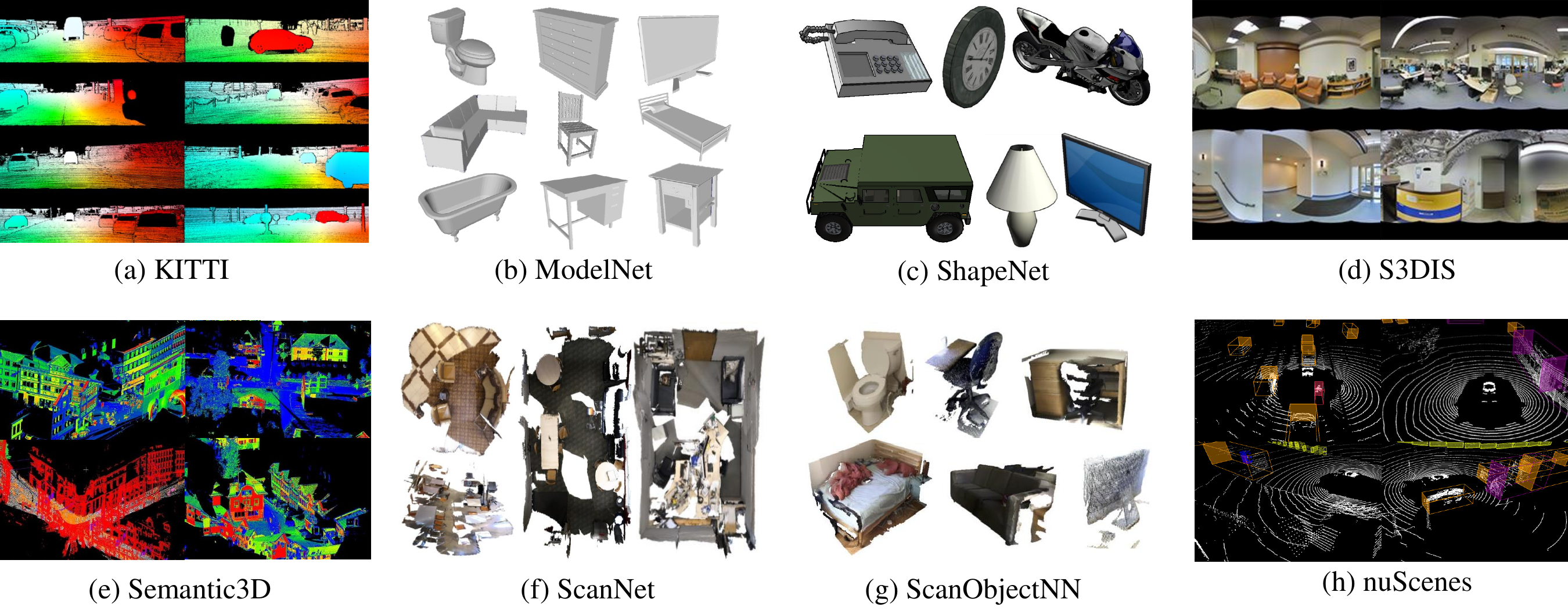}
    \caption{Most widely used 3D benchmark datasets for classification and segmentation.}
    \label{fig:point-cloud-dataset-samples}
    \Description{point-cloud-dataset-samples}
\end{figure}

\subsubsection{ModelNet (2015) \cite{Wu153DShapeNets}}
\label{subsubsec:modelnet}
A large-scale dataset of synthetic 3D shapes featuring 12,311 CAD-generated meshes from 40 types: chair, table, sofa, bed, bookshelf, lamp, car, airplane, bathtub, TV, etc. It serves as a foundational benchmark for evaluating 3D shape classification, recognition, and retrieval algorithms. ModelNet-C (2022) \cite{sun22ModelNet40C} is a derivative of the original dataset, which contains corrupted versions of the models, allowing for robustness testing.

\subsubsection{ShapeNet (2015) \cite{Chang15Shapenet}}
\label{subsubsec:shapenet}
Extends the scope of CAD-based datasets, offering over 50k CAD models (with surface color) and 25 rendered images per model from 55 semantic categories. In addition to supporting all the tasks from ModelNet, the dataset is also used to facilitate point cloud completion, reconstruction, and point cloud generation. There exist some variants of the core dataset, such as ShapeNetSem (2015) \cite{savva15ShapenetSem}- richer semantic labels with physical properties, ShapeNetPart (2016) \cite{Yi16ShapeNetPart}- includes part-based annotation, and ShapeNet-ViPC (2021) \cite{zhang21ShapenetViPC}- benchmarks point cloud completion.

\subsubsection{SUN RGB-D (2015) \cite{Song15SUNRGBD}}
\label{subsubsec:sunrgbd}
Accumulates 10k indoor RGB-D images from NYU-Depth V2 (2012) \cite{Silberman12NYUDepthV2}, Berkeley B3DO (2013) \cite{Janoch13BerkeleyB3DO}, and SUN3D (2013) \cite{Xiao13SUN3D}, and offers richer annotation supporting tasks like: scene categorization, object detection, object orientation, semantic segmentation, depth prediction, and room layout estimation. The dataset contains 37 object kinds from 47 scenes: bookshelf, bed, chair, table, door, picture, toilet, sink, bathtub, TV, etc.

\subsubsection{S3DIS (2016) \cite{Armeni16S3DIS}}
\label{subsubsec:s3dis}
Abbreviated from Stanford Large-Scale 3D Indoor Spaces, it contains scans of six large-scale indoor areas comprising 271 rooms, collectively covering over 6k square meters. It has 215M colored points labeled into 13 semantic classes: wall, ceiling, floor, column, beam, door, window, board, bookcase, chair, table, sofa, and clutter. Captured with a terrestrial LiDAR scanner, the dataset includes dense point-wise annotations for semantic, instance (later variant), and panoptic segmentation.

\subsubsection{Semantic3D (2017) \cite{Hackel17Semantic3D}}
\label{subsubsec:semantic3d}
A large-scale benchmark for point-level semantic segmentation of 30 real-world outdoor scenes. The data contains urban, suburban, and rural scenes, and is annotated into 8 classes: high vegetation, low vegetation, man-made terrain, natural terrain, cars, buildings, hard scape, and scanning artifacts. The point cloud data is generated by static terrestrial LiDAR sensors and has high density.

\subsubsection{ScanNet (2017) \cite{Dai17Scannet}}
\label{subsubsec:scannet}
Consists of over 15k 3D scans of annotated indoor scenes captured with consumer-grade RGB-D cameras. The dataset presents real-world challenges such as noise and occlusion, and supports a range of tasks, including classification, detection, segmentation, and depth estimation. ScanNet v2 is released later with improved support. The dataset is further used in conjunction with CAD datasets such as ShapeNet and ModelNet, such as Scan2CAD (2019) \cite{avetisyan19scan2cad}, for benchmarking tasks like CAD retrieval, pose estimation, and completion. \new{ScanNet200 \cite{Rozenberszki22ScanNet200} extends on ScanNet, providing annotations for 200 semantic classes. Scannet++ (2023) \cite{Yeshwanth23ScannetPP} borrows the idea from ScanNet and presents a dataset with higher-resolution 3D geometry and fine-grained semantic annotations, making it particularly suitable for tasks involving small or detailed objects.}

\subsubsection{Matterport3D (2017) \cite{Chang17Matterport3D}}
\label{subsubsec:matterport3d}
Indoor RGB-D dataset released by Princeton, Stanford, and Matterport Inc. It has 10k panoramic views constructed from 200k RGB-D images, which are annotated into 40 classes (furniture, bath fixtures, electronics, etc.) for benchmarks such as 3D semantic segmentation, navigation, and depth estimation.

\subsubsection{Pix3D (2018) \cite{Sun18Pix3d}}
\label{subsubsec:pix3d}
Curated for 3D shape understanding from 2D images featuring 6DoF pose information for each object along with respective CAD objects, segmentation masks, and camera viewpoints. There are 395 CAD models of IKEA furniture from 9 categories paired with 10k real-world images, making it valuable for shape retrieval, viewpoint estimation, and 3D reconstruction.

\subsubsection{ScanObjectNN (2019) \cite{Uy19ScanObjectNN}}
\label{subsubsec:scanobjectnn}
Focuses on real-world 3D object classification and recognition, unlike synthetic datasets like ModelNet \cite{Wu153DShapeNets} or ShapeNet \cite{Chang15Shapenet}, featuring around 15k object instances of 2902 3D unique models across 15 categories. The dataset is derived from filtering and annotating cropped objects from ScanNet \cite{Dai17Scannet} and SceneNN \cite{Binh16SceneNN}, and contains scene-level context like background and occlusions.

\subsubsection{nuScenes (2020) \cite{Caesar20nuScenes}}
\label{subsubsec:nuscenes}
A large-scale autonomous driving dataset, containing 1k driving scenes recorded in urban environments across Singapore and Boston. It comprises approximately 1.4M synchronized sensor frames captured by a sensor suite of 6 cameras, 5 radars, and 1 LiDAR scanner. It provides annotation for 23 object classes and supports tasks such as object detection, tracking, trajectory prediction, segmentation, and lane detection.

\subsubsection{Waymo Open Dataset (2020) \cite{Sun20WaymoOpenDataset} \cite{Ettinger21WaymoOpenMotionDataset}}
\new{A continuous effort by Waymo to support research in autonomous driving. The initial release of the dataset is called the Perception \cite{Sun20WaymoOpenDataset} dataset which consists 2,030 20-second segments captured at 10 Hz (390k frames) with synchronized multi-sensor data, including 5 LiDAR and 5 cameras. It provides 2D/3D bounding boxes, semantic and panoptic segmentation labels, key-point annotations, and map data. Then they released the Motion \cite{Ettinger21WaymoOpenMotionDataset} dataset that comprises 103k segments of 20-second sequences (20M frames) supporting motion forecasting research for autonomous driving. It includes 10.8 million tracked objects with 3D bounding boxes across vehicles, pedestrians, and cyclists, mined for behavior prediction scenarios such as unprotected turns, merges, lane changes, etc. The End-to-End Driving dataset offers 5k 20-second driving segments containing past and future ego-trajectories, vehicle state information, and text-based routing instructions. The dataset facilitates models to learn mappings from raw sensor observations to driving actions in realistic and complex environments.}

\subsubsection{Clear Pose (2022) \cite{Avidan22ClearPose}}
\label{subsubsec:clearpose}
A real-world dataset on monocular 3D understanding for transparent objects. It comprises 3D models and corresponding RGB images for 63 categories, encompassing both household and laboratory items. The dataset includes 3D pose annotations and depth maps, benchmarking single-image depth completion and object pose estimation.


\subsubsection{Other Datasets}
\label{subsubsec:other-datasets}
Figure \ref{fig:point-cloud-dataset-samples} presents samples from popular benchmark datasets. There are several other mainstream 3D benchmark datasets: \new{Replica (2019) \cite{Straub19Replica}- provides 18 high-resolution reconstructions of diverse indoor spaces, offering glass and mirror metadata, and planar, semantic, and instance labels; Campus3D (2020) \cite{Li20Campus3D}- an outdoor scan of the 1.58 km² National University of Singapore campus, captured via UAV photogrammetry and annotated into 2,530 instances across 24 classes.}
; IntrA (2020) \cite{yang2020intra}- 2,025 3D models of vessel segments to diagnose cerebral aneurysm; ARKitScenes (2021) \cite{dehghan21arkitscenes}- 5k scans of 1.6k scenes benchmarking object detection and depth upsampling; \new{Hypersim (2021) \cite{Roberts21Hypersim}- a large-scale, photorealistic synthetic indoor dataset, containing 77k images across 461 scenes designed by artists with rich per-pixel annotations}; MultiScan (2022) \cite{mao22multiscan}- uses commodity RGBD sensor to scan 117 indoor scenes with 10k objects for instance segmentation and part mobility estimation task; Objaverse (2022) \cite{Deitke22Objaverse}- 10M objects annotated by captions and tags ranging applications from segmentation to generation; SynLiDAR (2022) \cite{Xiao22SynLiDAR}- a synthetic LiDAR dataset with 20k scans from 32 classes for segmentation; OmniObject3D (2023) \cite{Wu23OmniObject3d}- 6k scanned objects with texture, multiview images, and videos in 190 categories; SceneFun3D (2024) \cite{delitzas24scenefun3d}- real-world indoor scene with 15k interaction annotations to encourage research on functionality segmentation and motion estimation; \new{WHU-Railway3D (2024) \cite{Qiu24WhuRailway3D}- semantic segmentation dataset covering 30 km of railway environments with 11 semantic categories; Belhouse3D (2025) \cite{Kumar25Belhouse3D}- synthetic dataset constructed from real-world references of 32 Belgian houses, simulating occlusions in real-world semantic segmentation scenarios}; etc.

\end{document}